\documentclass{article}

\usepackage{microtype}
\usepackage{graphicx}
\usepackage{subfigure}
\usepackage{booktabs} 

\usepackage{hyperref}

 \usepackage[accepted]{icml2023}

\usepackage{amsmath}
\usepackage{amssymb}
\usepackage{mathtools}
\usepackage{amsthm}

\usepackage[capitalize,noabbrev]{cleveref}

\usepackage{graphicx,psfrag,epsf,color}
\usepackage{enumerate}
\usepackage{natbib}
\setcitestyle{citesep={;}} 
\usepackage{multirow}
\usepackage{array}
\usepackage{bm}
\usepackage{float} 
\usepackage{url}

\usepackage{ulem}
\usepackage{amsmath,amsfonts,amsthm,amssymb,amsopn,bm}
\usepackage{multirow}
\usepackage{caption}
\usepackage{amssymb}
\usepackage{caption}

\usepackage[utf8]{inputenc} 
\usepackage[T1]{fontenc}    
\usepackage{url}            
\usepackage{amsfonts}       
\usepackage{nicefrac}       
\usepackage{microtype}      
\usepackage{xcolor}         
\usepackage{algorithm}
\usepackage{algorithmic}
\usepackage{authblk}
\usepackage{breqn}
\usepackage{microtype}
\usepackage{graphicx}
\usepackage{subfigure}
\usepackage{booktabs} 
\usepackage{mathtools}

\usepackage{bbding}
\theoremstyle{plain}
\newtheorem{theorem}{Theorem}[section]

\theoremstyle{definition}
\newtheorem{definition}[theorem]{Definition}
\newtheorem{assumption}[theorem]{Assumption}
\theoremstyle{remark}
\newtheorem{remark}[theorem]{Remark}

\newcommand\cruline{\bgroup\color{red}\markoverwith
{\textcolor{black}{\rule[-0.5ex]{2pt}{0.4pt}}}\ULon}
\newcommand\vibuline{\bgroup\color{blue}\markoverwith
{\textcolor{black}{\rule[-0.5ex]{2pt}{0.4pt}}}\ULon}
\newcommand\rnpuline{\bgroup\color{green}\markoverwith
{\textcolor{black}{\rule[-0.5ex]{2pt}{0.4pt}}}\ULon}
\newcommand\fruline{\bgroup\color{yellow}\markoverwith
{\textcolor{black}{\rule[-0.5ex]{2pt}{0.4pt}}}\ULon}

\usepackage{xparse,soul}

\ExplSyntaxOn
\NewDocumentCommand{\redulns}{m}
 {
  \seq_set_split:Nnn \l_tmpa_seq { ~ } { #1 }
  \seq_map_inline:Nn \l_tmpa_seq {\bgroup\color{red}\markoverwith
{\textcolor{black}{\rule[-0.5ex]{2pt}{0.4pt}}}\ULon{##1}~ } \unskip
 }
\NewDocumentCommand{\blackulns}{m}
 {
  \seq_set_split:Nnn \l_tmpa_seq { ~ } { #1 }
  \seq_map_inline:Nn \l_tmpa_seq {\bgroup\color{black}\markoverwith
{\textcolor{black}{\rule[-0.5ex]{2pt}{0.4pt}}}\ULon{##1}~ } \unskip
 }
\NewDocumentCommand{\blueulns}{m}
 {
  \seq_set_split:Nnn \l_tmpa_seq { ~ } { #1 }
  \seq_map_inline:Nn \l_tmpa_seq {\bgroup\color{blue}\markoverwith
{\textcolor{black}{\rule[-0.5ex]{2pt}{0.4pt}}}\ULon{##1}~ } \unskip
 }

 \NewDocumentCommand{\orangeulns}{m}
 {
  \seq_set_split:Nnn \l_tmpa_seq { ~ } { #1 }
  \seq_map_inline:Nn \l_tmpa_seq {\bgroup\color{orange}\markoverwith
{\textcolor{black}{\rule[-0.5ex]{2pt}{0.4pt}}}\ULon{##1}~ } \unskip
 }
  \NewDocumentCommand{\greenulns}{m}
 {
  \seq_set_split:Nnn \l_tmpa_seq { ~ } { #1 }
  \seq_map_inline:Nn \l_tmpa_seq {\bgroup\color{green}\markoverwith
{\textcolor{black}{\rule[-0.5ex]{2pt}{0.4pt}}}\ULon{##1}~ } \unskip
 }

\newcommand{\algorithmicreturn}{\textbf{Output:}}
\newcommand{\Out}{\item[\algorithmicreturn]}

\ExplSyntaxOff

\icmltitlerunning{Towards Trustworthy Explanation: On Causal Rationalization}

\begin{document}


\twocolumn[
\icmltitle{
           Towards Trustworthy Explanation: On Causal Rationalization}




\begin{icmlauthorlist}
\icmlauthor{Wenbo Zhang}{uci}
\icmlauthor{Tong Wu}{iqvia}
\icmlauthor{Yunlong Wang}{iqvia}
\icmlauthor{Yong Cai}{iqvia}
\icmlauthor{Hengrui Cai}{uci}
\end{icmlauthorlist}

\icmlaffiliation{uci}{Department of Statistics, University of California Irvine, California, USA}
\icmlaffiliation{iqvia}{Advanced Analytics, IQVIA,  Pennsylvania, USA}

\icmlcorrespondingauthor{Hengrui Cai}{hengrc1@uci.edu}



\icmlkeywords{Machine Learning, ICML}

\vskip 0.3in
]



\printAffiliationsAndNotice{}  

\begin{abstract}
With recent advances in natural language processing, rationalization becomes an essential self-explaining diagram to disentangle the black box by selecting a subset of input texts to account for the major variation in prediction. Yet, existing association-based approaches on rationalization cannot identify true rationales when two or more snippets are highly inter-correlated and thus provide a similar contribution to prediction accuracy, so-called spuriousness. To address this limitation, we novelly leverage two causal desiderata, non-spuriousness and efficiency, into rationalization from the causal inference perspective. We formally define a series of probabilities of causation based on a newly proposed structural causal model of rationalization, with its theoretical identification established as the main component of learning necessary and sufficient rationales. The superior performance of the proposed causal rationalization is demonstrated on real-world review and medical datasets with extensive experiments compared to state-of-the-art methods.

\end{abstract}

\section{Introduction}

Recent advancements in large language models have drawn increasing attention and have been widely used in extensive Natural Language Processing (NLP) tasks \citep[see e.g.,][]{NIPS2017_3f5ee243,kenton2019bert,lewis2019bart,brown2020language}. 
Although those deep learning-based models could provide incredibly outstanding prediction performance, it remains a daunting task in finding \textit{trustworthy explanations} to interpret these models' behavior, which is particularly critical in high-stakes applications in various fields. 
In healthcare, the use of electronic health records (EHRs) with raw texts is increasingly common to
forecast patients’ disease progression and assist clinicians in making decisions. These raw texts
serve as abstracts or milestones of a patient’s medical journey and characterize the
patient’s medical conditions \citep{estiri2021high,wu2021oa}. 
Beyond simply predicting clinical outcomes, doctors are more interested in understanding the decision-making process of predictive models thereby building trust, as well as extracting clinically meaningful and relevant insights \citep{liu2015temporal}. 
Discovering faithful text information hence is particularly crucial for improving the early diagnostic of severe disease and 
for making efficient clinical decisions.

\begin{figure*}[!htbp]
        \centering
            \vskip -0.1in
            \includegraphics[width=\textwidth]{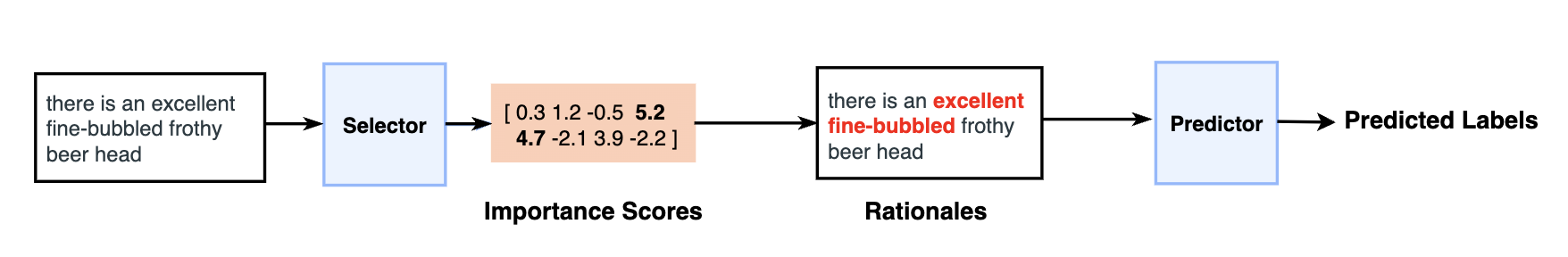}
            \vskip -0.2in
            \caption{This figure shows a standard selective rationalization framework for a beer review, which can be seen as a select-predict pipeline where firstly rationales are selected and then fed into the predictor.} 
            \vskip -0.15in\label{rationale}
\end{figure*}

Disentangling the black box in deep NLP models, however, is a notoriously challenging task \citep{alvarez2018towards}. 
There are a lot of research works focusing on providing trustworthy explanations for models, generally classified into post hoc techniques and self-explaining models \citep{danilevsky2020survey,rajagopal2021selfexplain}. To provide better model interpretation, self-explaining models are of greater interest, and selective rationalization is one popular type of such a model by highlighting important tokens among input texts  \citep[see e.g.,][]{deyoung2020eraser,paranjape2020information,jain2020learning,antognini2021multi,vafa2021rationales,chan2022unirex}. The general framework of selective rationalization as shown in \cref{rationale} consists of two components, a selector and a predictor. Those selected tokens by the trained selector are called \textit{rationales} and they are required to provide similar prediction accuracy as the full input text based on the trained predictor. Besides, the selected rationales should reflect the model's true reasoning process (faithfulness) \citep{deyoung2020eraser,jain2020learning} and provide a convenient explanation to people (plausibility) \citep{deyoung2020eraser,chan2022unirex}. 
Most of the existing works \citep{lei2016rationalizing,bastings2019interpretable,paranjape2020information} 
found rationales by maximizing the prediction accuracy for the outcome of interest 
based on input texts, and thus are association-based models. 

 \begin{figure}[ht]
            \vskip -0.05in\centering\includegraphics[width=0.65\columnwidth]{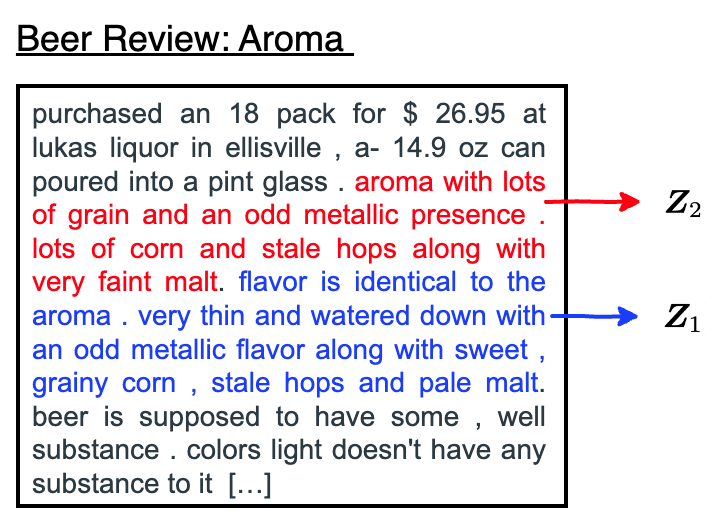}
            \vskip -0.1in
        \caption{Motivating example from the Beer review data. Red highlights the text related to the aroma (denoted as $\bm{Z}_2$) and blue highlights the text related to the palate (denoted as $\bm{Z}_1$).  The texts of aroma and palate are highly correlated with each other,  which makes them indistinguishable in terms of predicting the sentiment of interest.
        }   \label{example}
        \vskip -0.1in
\end{figure}

\begin{figure*}[!htb]
            \centering
            \vskip -0.1in\includegraphics[width=0.9\textwidth]{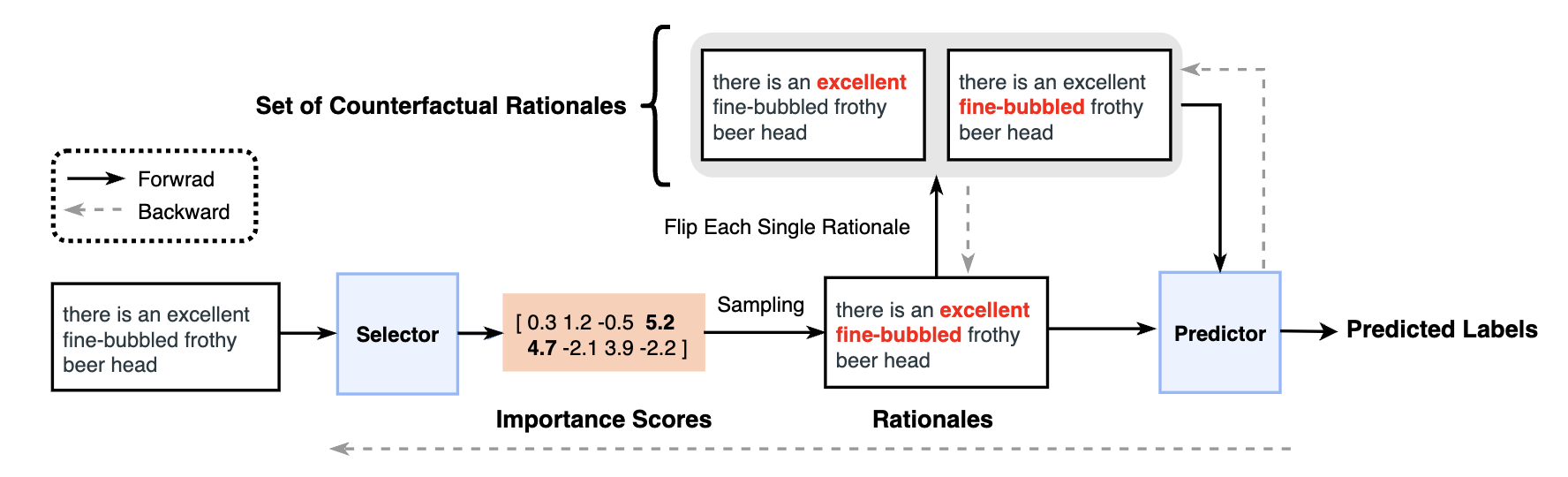}
    \vspace{-0.15in}\caption{The framework of the proposed causal rationalization. 
            Compared with the traditional selective rationalization in \cref{rationale},  our method adds the causal component (highlighted in gray) to generate counterfactual rationales.}
            \label{model}
            \vskip -0.1in
\end{figure*}


The major limitation of these {association-based} works (also see related works in Section \ref{related}) lies in falsely discovering \textit{spurious} rationales that may be related to the outcome of interest but do not indeed cause the outcome. Specifically, when two or more snippets are highly intercorrelated and provide a similar high contribution to prediction accuracy, the association-based methods cannot identify the true rationales among them. 
Here, the true rationales, formally defined in Section \ref{sec:frame} as the \textit{causal rationales}, are the true sufficient rationales to fully predict and explain the outcome without spurious information. As one example shown in Figure \ref{example}, one chunk of beer review comments covers two aspects, aroma, and palate. The reviewer has a negative sentiment toward the beer due to the unpleasant aroma 
(highlighted in red in \cref{example}). While texts in terms of the palate as highlighted in blue can provide a similar prediction accuracy 
as it is highly correlated with 
the aroma but not cause the sentiment of interest. Thus, these two selected snippets can't be distinguished purely based on the association with the outcome when they all have high predictive power. 
Such troublesome spuriousness  can be introduced in training as well owing to over-fitting to mislead the selector 
\citep{chang2020invariant}.  The predictor relying on these spurious features fails to achieve high \textit{generalization} performance when there is a large discrepancy between the training and testing data distributions 
\citep{scholkopf2021toward}.

In this paper, we propose a novel approach called causal rationalization aiming to find trustworthy explanations for general NLP tasks. Beyond selecting rationales based on purely optimizing prediction performance, our goal is to identify rationales with causal meanings. To achieve this goal, we introduce a novel concept as causal rationales by considering two causal desiderata \citep{wang2021desiderata}: non-spuriousness and efficiency. 
Here, \textit{non-spuriousness} means the selected rationales can capture features causally determining the outcome, and \textit{efficiency} means only essential and no redundant features are chosen. Towards these causal desiderata, our main \textbf{contributions} are threefold.

\vspace{0.05cm}

$\bullet$ We first formally define a series of the probabilities of causation (POC) for rationales accounting for non-spuriousness and efficiency at different levels of language, based on a newly proposed structural causal model of rationalization.\\
$\bullet$ We systematically establish the theoretical results for identifications of the defined POC of rationalization at the individual token level - 
the conditional probability of necessity and sufficiency ($\operatorname{CPNS}$) and derive the lower bound of $\operatorname{CPNS}$ under the relaxed identification assumptions for practical usage.\\
    $\bullet$ To learn necessary and sufficient rationales, we propose a novel algorithm that utilizes the lower bound of $\operatorname{CPNS}$ as the criteria to select the causal rationales. More specifically, we add the lower bound as a causality constraint into the objective function and optimize the model in an end-to-end fashion, as shown in \cref{model}.

    With extensive experiments, the superior performance of our causality-based rationalization method is demonstrated in the NLP dataset under both out-of-distribution and in-distribution settings.  
    The practical usefulness of our approach to providing trustworthy explanations for NLP tasks is demonstrated on
real-world review and EHR datasets.


\vspace{-0.1cm}
\subsection{Related Work}
\label{related}
\vspace{-0.05cm}
\noindent \textbf{Selective Rationalization.}
{Selective rationalization was firstly introduced in \citet{lei2016rationalizing} and now becomes an important model for interpretability, especially in the NLP domain \citep{bao2018deriving,paranjape2020information,jain2020learning,guerreiro2021spectra,vafa2021rationales,antognini2021rationalization}.  } To name a few recent developments, \citet{yu2019rethinking} proposed an introspective model 
under a cooperative setting with a selector and a predictor. \citet{chang2019game} extended their model to extract class-wise explanations. During the training phase, the initial design of the framework was not end-to-end because the sampling from these selectors was not differentiable. To address this issue, some later works adopted differentiable sampling, like Gumbel-Softmax or other reparameterization tricks \citep[see e.g.,][]{bastings2019interpretable,geng2020does,sha2021learning}. 
To explicitly control the sparsity of the selected rationales, \citet{paranjape2020information} derived a sparsity-inducing objective by using the information bottleneck. 
Recently, \citet{liu2022fr} developed a unified encoder to induce a better predictor by accessing  valuable information blocked by the selector. 


There are few recent studies that explored the issue of spuriousness in rationalization from the perspective of causal inference. \citet{chang2020invariant} proposed invariant causal rationales to avoid spurious correlation by considering data from multiple environments. \citet{plyler2021making} utilized counterfactuals produced in an unsupervised fashion using class-dependent generative models to address spuriousness. \citet{yue2023interventional} adopted a backdoor adjustment method to remove the spurious correlations in input and rationales. Our approach is notably different from the above methods. We only use a single environment, as opposed to multiple environments in \citet{chang2020invariant}. This makes our method more applicable in real-world scenarios, where collecting data from different environments can be challenging. Additionally, our proposed CPNS regularizer offers a different perspective on handling spuriousness, leading to improved generalization in out-of-distribution scenarios. Our work differs from  \citet{plyler2021making} and \citet{yue2023interventional} in that we focus on developing a regularizer (CPNS) to minimize spurious associations directly. This allows for a more straightforward and interpretable approach to the problem, which can be easily integrated into existing models.


\noindent \textbf{Explainable Artificial Intelligence (XAI).}
Sufficiency and necessity can be regarded as the fundamentals of XAI since they are the building blocks of all successful explanations \cite{watson2021local}. Recently, many researchers have started to incorporate these properties into their models. \citet{ribeiro2018anchors} proposed 
to find features that are sufficient to preserve current classification results. \citet{dhurandhar2018explanations} developed an autoencoder framework to find pertinent positive (sufficient) and pertinent negative (non-necessary) features which can preserve the current results. \citet{zhang2018interpreting} considered an approach to explain a neural network by generating minimal, stable, and symbolic corrections to change model outputs. Yet, sufficiency and necessity shown in the above methods are not defined from a causal perspective. Though there are a few works \citep{joshi2022all,balkir-etal-2022-necessity,galhotra2021explaining,watson2021local,beckers2021causal} defining these two properties 
with causal interpretations, all these works focus on post hoc analysis rather than a new model developed for rationalization. To the best of our knowledge, \citet{wang2021desiderata} is the most relevant work that quantified sufficiency and necessity for high-dimensional representations by extending  Pearl's POC \citep{pearl2000causality} and 
utilized those causal-inspired constraints 
to obtain a low-dimensional non-spurious and efficient representation. However, their method primarily assumed that the input variables are independent which allows a convenient estimation of their proposed POC. In our work, we generalize these causal concepts to rationalization and propose a more computationally efficient learning algorithm with relaxed assumptions. 

\begin{figure}[t]
        \centering
         \vskip -0.1in\includegraphics[width=0.18\textwidth]{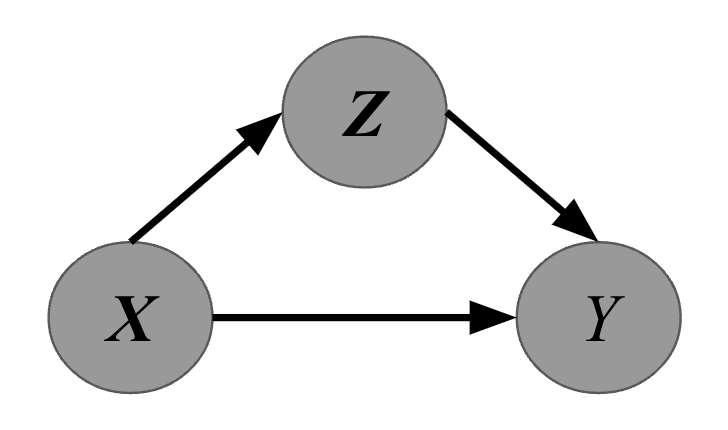}
         \vskip -0.1in
        \caption{Causal diagram of rationalization. It describes the data-generating process for the text $\boldsymbol{X}$, the true rationales $\boldsymbol{Z}$, and the label $Y$. Solid arrows denote causal relationships.}
        \label{causal diagram}
        \vskip -0.15in
\end{figure} 

 \begin{figure*}[t]
\vskip -0.1in
\centering
        \begin{subfigure}
        \centering
             \includegraphics[width=0.95\columnwidth]{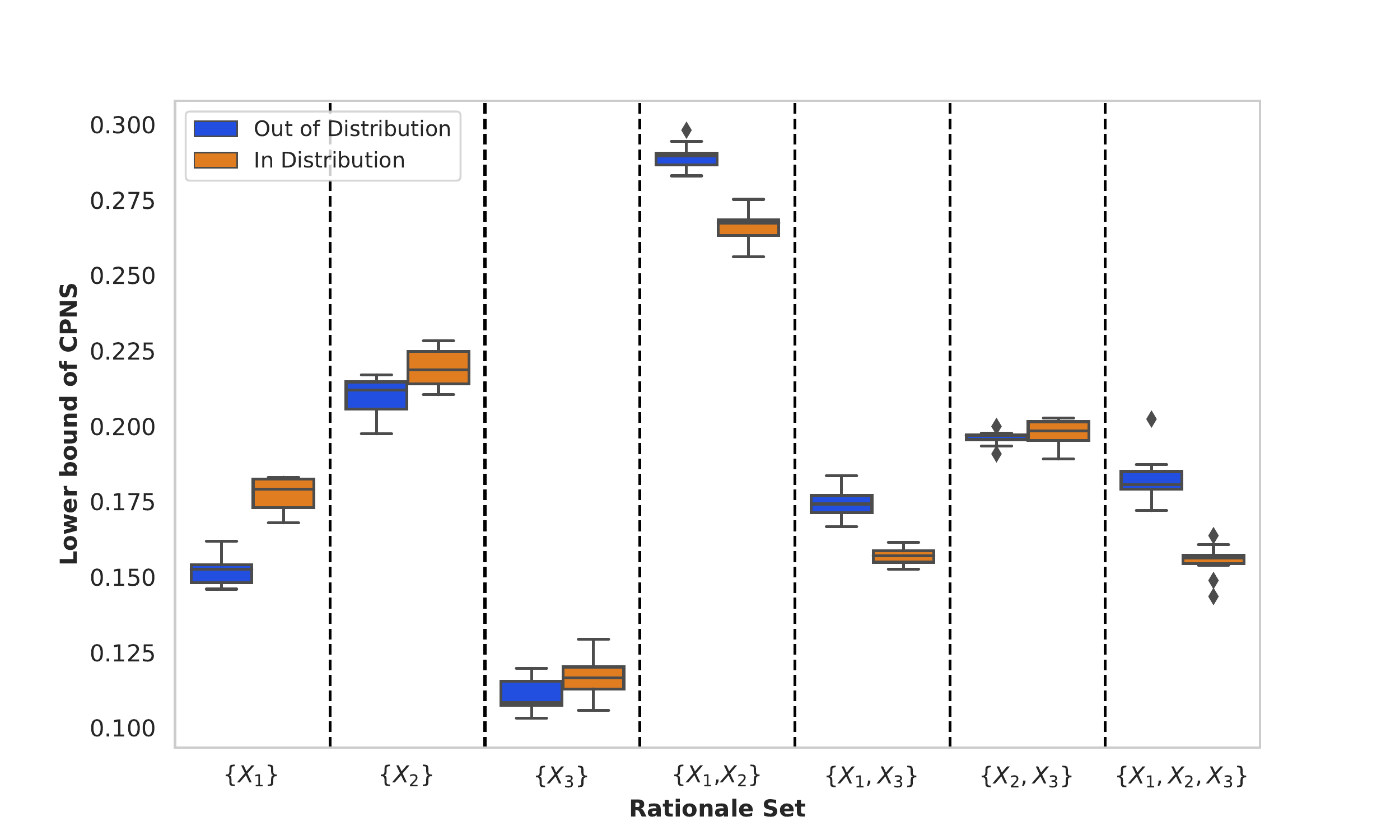}
        \end{subfigure}
        \begin{subfigure}
        \centering
\includegraphics[width=0.95\columnwidth]{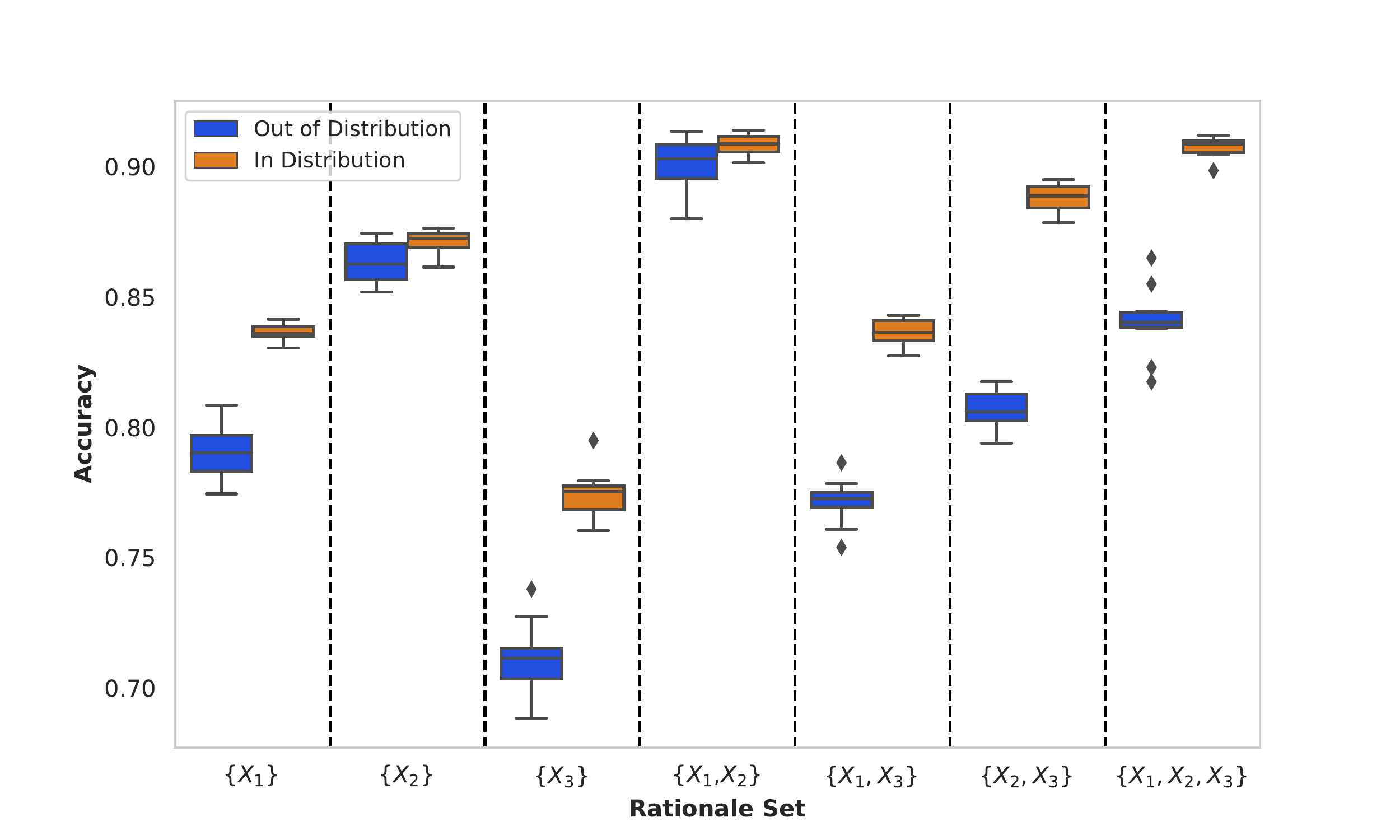}  
          \end{subfigure}  
        \vskip -0.2in
        \caption{We evaluate the average lower bound of $\operatorname{CPNS}$ and the accuracy of simulated data under the in-distribution (ID) and the out-of-distribution (OOD) setting. (a) ID: True rationale set $\{X_1,X_2\}$ achieves the most accurate forecasting as the set $\{X_1,X_2,X_3\}$ with all variables but its highest $\operatorname{CPNS}$ score can help distinguish it from others. (b) OOD: $\{X_1,X_2\}$ doesn't only provide the most accurate predictions but also has the highest lower bound $\operatorname{CPNS}$ value among all the possible rationale sets, meaning that using $\operatorname{CPNS}$ can select the true rationale to achieve OOD generalization.}
        \label{simu_plots}
        \vskip -0.15in
\end{figure*}

 \vspace{-0.2cm}
\section{Framework}\label{sec:frame}
\vspace{-0.1cm}

\textbf{Notations.} 
Denote $\boldsymbol{X}=\left(X_{1}, \cdots, X_d\right)$ as the input text with $d$ tokens, $\boldsymbol{Z}=\left(Z_{1}, \cdots, Z_d\right)$ as the corresponding selection where $Z_i \in \{0,1\}$ 
indicates whether the $i$-th token is selected or not by the selector, and $Y$ as the binary label of interest. Let $Y(\boldsymbol Z=\boldsymbol z)$ denote the potential value of $Y$ when setting $\boldsymbol Z$ as $\boldsymbol z$. Similarly, we define $Y(Z_i=z_i)$ as the potential outcome when setting the $Z_i$ as $z_i$ while keeping the rest of the selections unchanged. 
 

\vspace{-0.1cm}

\textbf{Structural Causal Model for Rationalization.}  A structural causal model (SCM) \citep{scholkopf2021toward} is defined by a causal diagram (where nodes are variables and edges represent causal relationships between variables) and modeling of the variables in the graph.  In this paper, we first propose an SCM for rationalization as follows with its causal diagram shown in Figure \ref{causal diagram}:
\vspace{-0.15cm}
\begin{eqnarray}\label{main_model}
\boldsymbol{X}=f(N_X),~~
\boldsymbol{Z}=g(\boldsymbol{X},N_Z),~~
Y=h(\boldsymbol{Z}\odot \boldsymbol{X},N_Y),
\end{eqnarray}
where 
$N_X,N_Y,N_Z$ are exogenous variables and $f,g,h$ are unknown functions that represent the causal mechanisms of $\boldsymbol{X},\boldsymbol{Z},Y$ respectively, with $\odot$ denoting the element-wise product. In this context, $g$ and $h$ can be regarded as the true selector and predictor respectively. Suppose we observe a data point with the text $\boldsymbol{X}$ and binary selections $\boldsymbol{Z}$, rationales can be represented by the event $\{\boldsymbol{X}_i\mathbb{I}(\boldsymbol{Z}_i=1)\}_{1\leq i \leq d}$, where $\mathbb{I}(\boldsymbol{Z}_i=1)$ indicates if the $i$-th token is selected, $\boldsymbol{X}_i$ is the corresponding text, and $d$ is the length of the text. 
\vspace{3mm}
\begin{remark}
The data generation process in \eqref{main_model} matches many graphical models in previous work \citep[see e.g.,][]{chen2018learning,paranjape2020information}. 
As a motivating example, consider the sentiment labeling process for the Beer review data. 
The labeler first locates all the important sub-sentences or words which encode sentiment information and marks them. After reading all the reviews, the labeler 
goes back to the previously marked text and makes the final judgment on sentiment. In this process, we can regard the first step of marking important locations of words as generating the selection of $\boldsymbol{Z}$ via reading texts $\boldsymbol{X}$. 
The second step is to combine the selected locations with raw text to generate rationales (equivalent to $\boldsymbol{Z}\odot \boldsymbol{X}$) and then the label $Y$ is generated through a complex decision function $h$. Discussions of potential dependences in \eqref{main_model} are provided in Appendix \ref{cla-scm}.
\end{remark}




\vspace{-0.2cm}
\section{Probability of Causation for Rationales}
\vspace{-0.1cm}
In this section, we formally establish a series of the probabilities of causation (POC) for rationales accounting for non-spuriousness and efficiency at different levels  of language, by extending 
\citet{pearl2000causality,wang2021desiderata}.

\begin{definition}\label{ps}
Probability of sufficiency ($\operatorname{PS}$) for rationales:
\begin{equation*}
\operatorname{PS} \triangleq P\left(Y(\boldsymbol{Z}=\boldsymbol{z})=y \mid \boldsymbol{Z} \neq \boldsymbol{z}, Y \neq y, \boldsymbol{X}= \boldsymbol{x} \right),
\end{equation*}
which indicates the sufficiency of rationales by evaluating the capacity of the rationales $\{\boldsymbol{X}_i\mathbb{I}(\boldsymbol{Z}_i=1)\}_{1\leq i \leq d}$ to “produce” the label 
if changing the selected rationales to the opposite. 
Spurious rationales shall have a low $\operatorname{PS}$. 
\end{definition}%


\begin{definition}
Probability of necessity ($\operatorname{PN}$) for rationales:
\begin{equation*}
\operatorname{PN} \triangleq  P\left( Y(\boldsymbol{Z}\neq \boldsymbol{z})\neq y \mid \boldsymbol{Z} = \boldsymbol{z},Y = y, \boldsymbol{X}= \boldsymbol{x} \right),
\label{pn}
\end{equation*}
which is the probability of rationales $\{\boldsymbol{X}_i\mathbb{I}(\boldsymbol{Z}_i=1)\}_{1\leq i \leq d}$ being a necessary cause of the label $\mathbb{I}\{Y=y\}$. Non-spurious rationales shall have a high $\operatorname{PN}$. 
\end{definition}
\vspace{-0.15cm}%
Desired true rationales should achieve 
non-spuriousness and efficiency simultaneously.  
This motivates us to define the probability of sufficiency and necessity as follows, with more explanations of these definitions in Appendix \ref{clarification}. 
\begin{definition}
Probability of necessity and sufficiency ($\operatorname{PNS}$) for rationales:
\begin{equation*}
\operatorname{PNS} \triangleq P\left(Y(\boldsymbol{Z} \neq \boldsymbol{z}) \neq y, Y(\boldsymbol{Z}=\boldsymbol{z})=y \mid \boldsymbol{X}=\boldsymbol{x} \right).
\label{pns}
\end{equation*}
\end{definition}%
\vspace{-0.2cm}

Here, $\operatorname{PNS}$ can be regarded as a good proxy of causality and we illustrate this in Appendix \ref{significance}. When both the selection $\boldsymbol{Z}$ and the label $Y$ are univariate binary and $\boldsymbol{X}$ is removed, our defined $\operatorname{PN}$, $\operatorname{PS}$, and $\operatorname{PNS}$ boil-down into the classical definitions of POC in Definitions 9.2.1 to 9.2.3 of \citet{pearl2000causality}, respectively. In addition, our definitions imply that the input texts are fixed and 
POC is mainly detected through interventions on selected rationales. This not only reflects the data generation process we proposed in Model \eqref{main_model} but also distinguishes our settings with existing works \citep[][]{pearl2000causality,wang2021desiderata,cai2023learning}. 
To serve the role of guiding rationale selection, we extend the above definition to 
a conditional version for a single rationale.
\begin{definition}\label{df_single_cpns}
Conditional probability of necessity and sufficiency ($\operatorname{CPNS}$) for the $j$th selection:
\begin{equation*}
\begin{aligned}
\operatorname{CPNS}_j \triangleq P(&Y\left(Z_{j}=z_{j}, \boldsymbol{Z}_{-j}=\boldsymbol{z}_{-j}\right)=y,\\
&Y\left(Z_{j} \neq z_{j}, \boldsymbol{Z}_{-j}=\boldsymbol{z}_{-j}\right) \neq y \mid \boldsymbol{X}=\boldsymbol{x} ).
\end{aligned}
\end{equation*}
\end{definition}%
Definition \ref{df_single_cpns} mainly focuses on a single rationale. The importance of Definitions \ref{ps}-\ref{df_single_cpns} can be found in Appendix \ref{significance}. We further define $\operatorname{CPNS}$ over all the selected rationales as follows. 
\begin{definition}\label{df_ocpns}
The $\operatorname{CPNS}$ over selected rationales: 
\begin{equation*}
\operatorname{CPNS} \triangleq \prod_{j \in \mathbf{r}} ( \operatorname{CPNS}_j)^{\frac{1}{|\mathbf{r}|} },
\end{equation*}
where  $\mathbf{r}$ is the index set for the selected rationales and $|\mathbf r|$ is the number of selected rationales.
\end{definition}
The proposed $\operatorname{CPNS}$ can be regarded as a geometric mean and we discuss why utilizing this formulation in Appendix \ref{geometric}, with more connections among Definitions \ref{df_single_cpns}-\ref{df_ocpns} in Appendix \ref{token-level}. To generalize $\operatorname{CPNS}$ to data, we can utilize the average $\operatorname{CPNS}$ over the dataset to represent an overall measurement. We denote the lower bound of $\operatorname{CPNS}$ and $\operatorname{CPNS}_j$ by $\underline{\operatorname{CPNS}}$ and $\underline{\operatorname{CPNS}}_j$, respectively, with their theoretical derivation and identification  in Section \ref{sec:lowerbound}.


\subsection{Example of Using POC to Find Causal Rationales} 

We utilize a toy example to demonstrate why $\mathrm{CPNS}$ is useful for in-distribution (ID) feature selection and out-of-distribution (OOD) generalization. Suppose there is a dataset of sequences, where each sequence can be represented as $\boldsymbol{X}=\left(X_{1}, \ldots,X_{l}\right)$ with an equal length as $l=3$ and a binary label $Y$. 
Here, we set $\{X_1,X_2\}$ are true rationales and $\{X_3\}$ is an irrelevant/spurious feature. 

The process of generating such a dataset is described below. Firstly, we generate rationale features $\{X_1,X_2\}$ following a bivariate normal distribution with positive correlations. To create spurious correlation, we generate irrelevant features $\{X_3\}$ by using mapping functions which map $\{X_1,X_2,\epsilon\}$ to $\{X_3\}$ where $\epsilon \sim N(0,0.5)$.  Here we use a linear mapping function $X_3=X_1+\epsilon$. Thus irrelevant features are highly correlated with $X_1$.

There are 3 simulated datasets: the training dataset, the in-distribution test data, and the out-of-distribution test data. The training data $\left\{\boldsymbol{x}_{i}^{\text {train }}\right\}_{i=1}^{n_{\text {train }}}$ and the in-distribution test data $\left\{\boldsymbol{x}_{i}^{\text {test-in }}\right\}_{i=1}^{n_{\text {test-in }}}$ follows the same generation process, but for out-of-distribution test data $\left\{\boldsymbol{x}_{i}^{\text {test-out }}\right\}_{i=1}^{n_{\text {test-out }}}$, we modify the covariance matrix of ${X_1,X_2}$ to create a different distribution of the features. Then we make the label $Y$ only depends on $\left(X_{1},X_{2}\right)$. This is equivalent to assuming that all the rationales are in the same position and the purpose is to simplify the explanations. For a single $\boldsymbol{x}_i$, we simulate $\mathrm P(y_i=1|\boldsymbol{x}_i)=\pi(\boldsymbol{x}_i)$  below:
\begin{equation*}
    \pi(\boldsymbol{x}_i)=\frac{1}{1+e^{-(\beta_0+\beta_1x_{i1}+\beta_2x_{i2})}}.
\end{equation*}
Then we use threshold value $0.5$ to categorize the data into one of two classes: $y_i = 1$ if $\pi \geq 0.5$ and $y_i = 0$ if $\pi \leq 0.5$. Since the dataset includes $3$ features, there are $6$ combinations of the rationale (we ignore the rationale containing no features). For $i$th rationale, we would fit a logistic regression model by using only selected features and refit a new logistic regression model with subset features to calculate $\underline{\operatorname{CPNS}}_i$. In our simulation, we set $n_{\text {train }}=n_{\text {test-in }}=n_{\text {test-out }}=2000$, $\beta_0=1$, $\beta_1=0.5$ and $\beta_2=1$. 

The results of the average of $\underline{\operatorname{CPNS}}_1$ and $\underline{\operatorname{CPNS}}_2$, and the accuracy measured in OOD and ID test datasets are shown in \cref{simu_plots} over $10$ replications. It can be seen that true rationales ($\{X_1,X_2\}$) yield the highest scores of the average $\underline{\operatorname{CPNS}}$ and accuracy in both OOD and ID settings. 
This motivates us to identify true rationales by maximizing the score of $\operatorname{CPNS}$ or the lower bound.

\vspace{-0.2cm}
\section{
Identifiability and Lower Bound of CPNS}\label{sec:lowerbound} 
As has been shown, the proposed $\operatorname{CPNS}$ helps to discover the true rationales that achieve high OOD generalization. Yet, due to the unobserved counterfactual events in the observational study, we need to identify $\operatorname{CPNS}$ 
as statistically estimated quantities. To this end, 
we generalize three common assumptions in causal inference  \citep{pearl2000causality,vanderweele2009conceptual,imai2010general} to rationalization, including consistency, ignorability, and monotonicity. 

\vspace{0.02cm}

\begin{assumption}
\begin{equation}
\vspace{-0.1cm}
 \bullet \text{ Consistency:  } \boldsymbol{Z}=\boldsymbol{z} {\rightarrow }Y(\boldsymbol{Z}=\boldsymbol{z}
 )=Y.
 \label{consistency}~~~~~~~~~~~~~~~~~~~~~~~~~~~
\end{equation}
$\bullet$ Ignorability:
\begin{equation}
\begin{aligned}
\{&Y(Z_j=z_j,\boldsymbol{Z}_{-j}=\boldsymbol{z}_{-j}),\\ 
&Y(Z_j\neq z_j,\boldsymbol{Z}_{-j}=\boldsymbol{z}_{-j})\} \perp \boldsymbol{Z} \mid \boldsymbol{X}.
\label{ignorability}
\end{aligned}
\end{equation}
$\bullet$ Monotonicity (for $\boldsymbol{X}=\boldsymbol{x}$, $Y$ is monotonic relative to $\boldsymbol{Z}_i$):
\begin{equation}
\begin{aligned}
&\left\{Y\left(Z_{j} \neq z_{j}, \boldsymbol{Z}_{-j}=\boldsymbol{z}_{-j}\right)=y\right\}\\
& \wedge\left\{Y\left(Z_{j}=z_{j}, \boldsymbol{Z}_{-j}=\boldsymbol{z}_{-j}\right) \neq y\right\}=\text { False },
\label{monotonicity}
\end{aligned}
\end{equation}
where $\wedge$ is the logical operation AND. For two events $A$ and $B$, $A\wedge B =$ True if $A=B=$ True, $A\wedge B=$ False otherwise.
\end{assumption}%


Here in the first assumption, the left-hand side means the observed selection of tokens to be $Z=z$, and the right-hand side means the actual label observed with the observed selection $Z=z$ is identical to the potential label we would have observed by setting the selection of tokens to be $Z=z$. The second assumption in causal inference usually means no unmeasured confounders, which is automatically satisfied under randomized trials. For observational studies, we rely on domain experts to include as many features as possible to guarantee this assumption. In rationalization, it means our text already contains all information. For monotonicity assumption, it indicates that a change on the wrong selection can not, under any circumstance, make $Y$ change to the true label. In other words, true selection can increase the likelihood of the true label. The theorem below shows the identification and the partial identification results for $\operatorname{CPNS}$.

\begin{theorem}
\label{theorem}
Assume the causal diagram in Figure \ref{causal diagram} holds.
$ \bullet$ If assumptions \eqref{consistency}, \eqref{ignorability}, and \eqref{monotonicity} hold, then $\operatorname{CPNS}_{j}$ can be identified by
\begin{equation}
\resizebox{0.95\hsize}{!}{$
\begin{aligned}
\operatorname{CPNS}_{j}=&P(Y=y \mid Z_{j}=z_{j}, \boldsymbol{Z}_{-j}=\boldsymbol{z}_{-j},\boldsymbol{X}= \boldsymbol{x})\\
&-P(Y=y \mid Z_{j} \neq z_{j}, \boldsymbol{Z}_{-j}=\boldsymbol{z}_{-j},\boldsymbol{X}= \boldsymbol{x}).
\nonumber
\end{aligned}$}
\end{equation}
$ \bullet$ If only assumptions \eqref{consistency} and \eqref{ignorability} hold, $\operatorname{CPNS}_{j}$ is not identifiable but its lower bound can be calculated by
\begin{equation}
\resizebox{1\hsize}{!}{$
\begin{aligned}
\underline{\operatorname{CPNS}}_{j}=&\max \Bigl[0,P(Y=y \mid Z_{j}=z_{j}, \boldsymbol{Z}_{-j}=\boldsymbol{z}_{-j},\boldsymbol{X}= \boldsymbol{x})\\
&-P(Y=y \mid Z_{j} \neq z_{j}, \boldsymbol{Z}_{-j}=\boldsymbol{z}_{-j},  \boldsymbol{X}= \boldsymbol{x})\Bigr].
\nonumber
\end{aligned}$}
\end{equation} 
\end{theorem} 
The detailed proof is provided in \cref{proof}. \cref{theorem} generalizes Theorem 9.2.14 and 9.2.10 of \citet{pearl2000causality} to multivariate binary variables. This is similar to the single binary variable case since for each single $Z_j$, the conditional event $\{\boldsymbol{Z}_{-j}=\boldsymbol{z}_{-j},\boldsymbol{X}= \boldsymbol{x}\}$ doesn't change. The first part of the theorem provides identification results for the counterfactual quantity $\operatorname{CPNS}_j$ and we can estimate it using observational data and the flipping operation as shown in \cref{model}. For example, given a piece of text $\boldsymbol{x}$, $P(Y=y \mid Z_{j}=z_{j}, \boldsymbol{Z}_{-j}=\boldsymbol{z}_{-j},\boldsymbol{X}= \boldsymbol{x})$ can be estimated by feeding the original rationales $\boldsymbol{z}$ produced by the selector 
to the predictor, 
and $P(Y=y \mid Z_{j}\neq z_{j}, \boldsymbol{Z}_{-j}=\boldsymbol{z}_{-j},\boldsymbol{X}= \boldsymbol{x})$ can be estimated by predicting the label of the counterfactual rationals which is obtained by flipping the $z_j$. We can notice that Theorem \ref{theorem} can be generalized when $z_i$ represents whether to mask a clause/sentence for finding clause/sentence-level rationales. 
Since the monotonicity assumption (\ref{monotonicity}) is not always satisfied, especially during the model training stage. Based on \cref{theorem}, we can relax the monotonicity assumption and derive the lower bound of $\operatorname{CPNS}_j$. Since a larger lower bound can imply higher $\operatorname{CPNS}_j$ but a larger upper bound doesn't, we focus on the lower bound here and utilize it as a substitution for $\operatorname{CPNS}_j$. Combining each rationale, we can get the lower bound of $\operatorname{CPNS}$ as $\underline{\operatorname{CPNS}} =\prod_{j \in \mathbf{r}} ( \underline{\operatorname{CPNS}_j)}^{\frac{1}{|\mathbf{r}|} }$.

\begin{algorithm}[!t]
\caption{Causal Rationalization}
\begin{algorithmic}
 \REQUIRE
 Training dataset $\mathcal{D}=\{(\boldsymbol{x}_i,y_i)\}_{i=1}^N$, parameters $k$, $\alpha$, $\mu$ and $\lambda$.
 

 \STATE \textbf{Begin:} Initialize the parameters of selector $g_\theta(\cdot)$ and predictor $h_\phi(\cdot)$, where $\theta$ and $\phi$ denote their parameters
 

 \WHILE{not converge} 
 \STATE  Sample a batch $\{(\boldsymbol{x}_i,y_i)\}_{i=1}^n$ from $\mathcal{D}$

\STATE Generate selections $\mathcal S=\{\boldsymbol{z}_i\}_{i=1}^n$ through Gumbel-Softmax sampling 

 \FOR{$i=1,\ldots ,n$}
 
  \STATE  Get a random sample $\mathbf{r}_i^{(k)}$ from index set $\mathbf{r}_i$ where $\mathbf{r}_i$ represents the set of rationales that are selected as 1 in $\boldsymbol{z}_i$ and its size equals $k\% \times \text{length}(\mathbf{x}_i)$ 
 
 \FOR{$j=1,\ldots,|\mathbf{r}_i^{(k)}|$}

 \STATE Generate counterfactual selections $\boldsymbol{z}_{i(j)}$ by flipping the $j$th index of the index set $\mathbf{r}_i^{(k)}$\;
 
\ENDFOR
\ENDFOR

\STATE Get a new batch of selections $\tilde{\mathcal{S}}=\{\boldsymbol{z}_{i(j)}\}_{j=1,\cdots,|\mathbf{r}_i^{(k)}|}^{i=1,\cdots,n}$ and set $\mathcal{S}_{\text{all}}=\mathcal{S}\bigcup \tilde{\mathcal{S}}$

\STATE Compute $\mathcal{L}$ via Eq(\ref{obj2}) by using $\mathcal{S}_{\text{all}}$ and $\mathcal{D}$

\STATE Update parameters $\theta=\theta-\alpha \nabla_{\theta} \mathcal{L}; \phi=\phi-\alpha \nabla_{\phi} \mathcal{L}$ 
\ENDWHILE

\Out selector $g_\theta(\cdot)$ and predictor $h_\phi(\cdot)$

\end{algorithmic}
\label{algo}
\end{algorithm}

\setlength{\textfloatsep}{15pt} 

\vspace{-0.2cm}
\section{Learning Necessary and Sufficient Rationale}
\vspace{-0.1cm}
In this section, we propose to learn necessary and sufficient rationales by incorporating $\underline{\operatorname{CPNS}}$ as the causality constraint into the objective function.

\vspace{-0.2cm}
\subsection{Learning Architecture}
\vspace{-0.1cm}
Our model framework consists of a selector $g_{\theta}(\cdot)$ and a predictor $h_{\phi}(\cdot)$ as standard in the traditional rationalization
approach, where $\theta$ and $\phi$ denote their parameters. We can get the selection $\boldsymbol{Z}=g_{\theta}(\boldsymbol{X})$ and fed it into predictor to get $Y=h_{\phi}(\boldsymbol{Z}\odot \boldsymbol{X})$ as shown in Figures \ref{model} and \ref{causal diagram}. One main difference between causal rationale and original rationale is that we generate a series of counterfactual selections by flipping each dimension of the selection $\bm{Z}$ we obtained from the selector. Then we feed raw rationales with new counterfactual rationales into our predictor to make predictions. Considering the considerable cost of obtaining reliable rationale annotations from humans, we only focus on unsupervised settings. Our goal is to make the selector select rationales 
with the property of necessity and sufficiency and our predictor can simultaneously provide accurate predictions given such rationales. 

\vspace{-0.2cm}
\subsection{Role of POC and Its Estimation}
\vspace{-0.1cm}
For the $j$-th token to be selected as a rationale, according to the results of Theorem \ref{theorem}, we expect $P(Y=y \mid Z_{j}=z_{j}, \boldsymbol{Z}_{-j}=\boldsymbol{z}_{-j},  \boldsymbol{X}= \boldsymbol{x})$ to be large while $\operatorname{P}(Y=y_\mid Z_{j} \neq z_{j}, \boldsymbol{Z}_{-j}=\boldsymbol{z}_{-j},  \boldsymbol{X}= \boldsymbol{x})$ to be small. Here, these two probabilities can be estimated from the predictor using selected rationales and flipped rationales, respectively, through the deep model. The empirical estimation of $\underline{\operatorname{CPNS}_j}$ is denoted as $\underline{\widehat{\mathrm{CPNS}}_j}$. The estimated lower bound is used as the causal constraint to reflect the necessity and sufficiency of a token of determining the outcome. If $\underline{\widehat{\mathrm{CPNS}}_j}$ is large, we expect the corresponding token to be selected into the final set of rationales.

\vspace{-0.2cm}
\subsection{Learning Necessary and Sufficient Rationales}
\vspace{-0.1cm}
Given a training dataset $\mathcal{D}$, we consider the following objective function utilizing lower bound of $\operatorname{CPNS}$ to train the causal rational model:
\vspace{-0.1cm}
\begin{equation}
\resizebox{1.05\hsize}{!}{$
\begin{aligned}
 \underset{\theta, \phi}{\min} \hspace{1mm} \mathcal{L}
= \underset{\theta, \phi}{\min} \hspace{1mm} \mathbb{E}_{(\bm{x},y)\sim \mathcal{D}} \Biggr[ &L\left(y, \widehat{y}\right)+\lambda \delta\left(\boldsymbol{z}\right) -\underbrace{\mu\sum_{j\in \mathbf{r}^{(k)}}\frac{\log\underline{ \widehat{\mathrm{CPNS}}^{+}_j}}{|\mathbf{r}^{(k)}|}}_{\text {Causality Constraint }}\Biggr],
\label{obj2}
\end{aligned}$}
\end{equation}
where $\widehat{y}=h_\phi\left(\boldsymbol{z} \odot \boldsymbol{x}\right)$, $\boldsymbol{z}=g_{\theta}(\boldsymbol{x})$, $L(\cdot,\cdot )$ defined as the cross-entropy loss, $\delta(\cdot)$ is the sparsity penalty to control sparseness of rationales, $\mathbf{r}_i^{(k)}$ denotes the a random subset with size equal $k\%$ of the sequence length, $\lambda$ and $\mu$ are the tuning parameters. The reason we sample a random subset $\mathbf{r}_i^{(k)}$ is due to the computational cost of flipping each selected rationale. To avoid  the negative infinity value of $\log\underline{ \widehat{\mathrm{CPNS}}^{+}_j}$ when the estimated $\underline{ \widehat{\mathrm{CPNS}}_j}$ is 0, we add a small constant $c$ to get $\underline{ \widehat{\mathrm{CPNS}}^{+}_j}=\underline{ \widehat{\mathrm{CPNS}}_j}+c$. Since the value of $c$ has no influence on the optimization of the objective function, we set $c=1$. Our proposed algorithm to solve (\ref{obj2}) is presented in \cref{algo}.  

\vspace{2mm}
\begin{remark}
One of our motivations is from medical data. In those data, important rationales are not necessarily consecutive medical records, and can very likely scatter all over patient longitudinal medical journeys. However, continuity is a natural property in text data. Therefore, we first conduct the experiments without the continuity constraint in the next section and then conduct experiments with the continuity constraint in Appendix \ref{add-cont}.
\end{remark}

\vspace{-0.2cm}
\section{Experiments and Results}
\subsection{Datasets}
\noindent \textbf{Beer Review Data.}
The Beer review dataset is a multi-aspect sentiment analysis dataset with sentence-level annotations \citep{mcauley2012learning}. Considering our approach focuses on the token-level selection and we don't use continuity constraint, we utilize the Beer dataset, with three aspects: appearance, aroma, and palate, collected from \citet{bao2018deriving} where token-level true rationales are given.

\noindent \textbf{Hotel Review Data.}
The Hotel Review dataset is a form of multi-aspect sentiment analysis from \citet{wang2010latent} and we mainly focus on the location aspect. 

\noindent \textbf{Geographic Atrophy (GA) Dataset}. 
The proprietary GA dataset used in this study includes the medical claim records who are diagnosed with Geographic Atrophy or have risk factors. Each claim records the date, the ICD-10 codes of the medical service, where the codes represent different medical conditions and diseases, and the description of the service. We are tasked to utilize the medical claim data to find high-risk GA patients and reveal important clinical indications using the rationalization framework. This dataset doesn't provide human annotations because it requires a huge amount of time and money to hire domain experts to annotate such a large dataset. 




\subsection{Baselines, Implementations, and Metrics}
\noindent \textbf{Baselines.}
We consider five baselines: rationalizing neural prediction (RNP), variational information bottleneck (VIB), folded rationalization (FR), attention to rationalization (A2R), and invariant rationalization (INVART). 
RNP is the first select-predict rationalization approach proposed by \citet{lei2016rationalizing}. VIB utilizes a discrete bottleneck objective to select the mask \citep{paranjape2020information}. FR doesn't follow a two-stage training for the generator and the predictor, instead, it utilizes a unified encoder to share information among the two components \citep{liu2022fr}. A2R \citep{yu2021understanding} combines both hard and soft selections to build a more robust generator. INVART \citep{chang2020invariant} enables the predictor to be optimal across different environments to reduce spurious selections.

\noindent \textbf{Implementations.}
We utilize the same sparse constraint in VIB, and thus the comparison between our method and VIB is also an ablation study to verify the usefulness of our causal module. For a fair comparison, all the methods don't include the continuous constraint. Following \citet{paranjape2020information}, we utilize BERT-base as the backbone of the selector and predictor for all the methods for a fair comparison. We set the hyperparameter $\mu=0.1$ and $k=5\%$ for the causality constraint. See more details in Appendix \ref{implt_appen}. 

\noindent \textbf{Metrics.}
For real data experiments of the Beer review dataset, we utilize prediction accuracy (Acc), Precision (P), Recall (R), and F1 score (F1), where P, R, and F1 are utilized to measure how selected rationales align with human annotations, and of the GA dataset, the area under curve (AUC) is used. For synthetic experiments of the Beer review dataset, we adopt Acc, P, R, F1, and False Discovery Rate (FDR) where FRD evaluates the percentage of injected noisy tokens captured by the model which have a spurious correlation with labels.

\setlength{\textfloatsep}{20.0pt plus 2.0pt minus 4.0pt}

\begin{table*}[t] 
\centering
\caption{Results on the Beer review dataset. Top-$10\%$ tokens are selected for the test datasets. Check updated results for FR in Remark \ref{update fr} }
\vskip -0.1in
\resizebox{2.05\columnwidth}{!}{%
\begin{tabular}{lcccccccccccc} 
\toprule
\multirow{2}{*}{ Methods } & \multicolumn{4}{c}{ Appearance } & \multicolumn{4}{c}{ Aroma } & \multicolumn{4}{c}{ Palate } \\
\cmidrule(lr){2-5}
\cmidrule(lr){6-9}
\cmidrule(lr){10-13}
& Acc $(\uparrow)$ & P $(\uparrow)$ &  R $(\uparrow)$ & F1 $(\uparrow)$ & Acc $(\uparrow)$  & P $(\uparrow)$ &R $(\uparrow)$ & F1 $(\uparrow)$ & Acc $(\uparrow)$  & P $(\uparrow)$& R $(\uparrow)$& F1 $(\uparrow)$ \\
\midrule
VIB & $92.2(1.1)$  &$42.7(2.1)$ &$21.6(1.9)$ & $26.8(2.0)$ & $83.5(1.5)$ &$43.4(1.1)$ &$25.6(2.2)$ & $30.1(1.0)$ & $74.3(2.4)$ &$27.6(1.4)$ & $24.1( 3.5)$ & $23.6(2.4)$  \\
RNP & $91.0(1.1)$ &$48.7(4.5)$ &$11.7(0,9)$ & $20.0(1.5)$ & $82.4(0.8)$ &$44.2(2.6)$ &$20.7(3.2)$ & $27.6(4.3)$ & $74.0(0.9)$ &$25.1(1.8)$ &$21.9(2.0)$ & $22.8(4.9)$  \\
FR & $\mathbf{94.5(0.6)}$  & $37.7(1.9)$&$19.1(0.8)$ & $23.7( 1.1)$ & $86.0( 1.6)$ &$40.3(2.5)$ &$22.3(3.5)$ & $28.0(1.9)$ & $\mathbf{77.0(1.9)}$ &$25.8(1.0)$ & $23.6(1.3)$& $22.4( 1.6)$  \\
A2R &$92.2(1.3)$& $\mathbf{49.1(1.0)}$&$18.9(1.6)$ &$25.9(0.9)$ &$82.5(1.8)$& $51.2(2.7)$ &$21.2(1.9)$ &$29.8(2.4)$ & $74.3(3.6)$ &$31.8(2.5)$& $24.3(2.1)$ &$25.4(2.0)$\\
INVRAT & $ 94.0(0.9)$ &$45.8(1.4)$ & $20.6(1.3)$ & $26.7(1.6)$&$84.3(2.0)$ & $43.0(4.5)$& $23.1(3.2)$ & $28.3(4.0)$ & $76.8(3.8)$ & $28.7(2.2)$ &$23.5(1.0)$& $23.8(2.0)$ \\ 
CR(ours)  & $93.1(1.1)$ &$45.3(1.7)$ &$\mathbf{22.0(1.1)}$ & $\mathbf{28.0(1.2)}$  &  $\mathbf{86.6(1.7)}$ &$\mathbf{60.3( 3.0)}$  &$\mathbf{35.4(2.1)}$& $\mathbf{39.0(4.1)}$&  $75.7(3.2)$ & $\mathbf{32.5(2.8)}$ & $\mathbf{25.9(0.5)}$& $\mathbf{26.5 (1.2)}$ \\
\bottomrule
\end{tabular}
}
\label{beer real}
\end{table*}

\begin{table*}[t] 
\centering
\caption{Results on the Spurious Beer review dataset. Top-$10\%$ tokens are highlighted for evaluation.  }
\vskip -0.1in
\resizebox{1.8\columnwidth}{!}{%
\begin{tabular}{lcccccccccc} 
\toprule
\multirow{2}{*}{ Methods }  & \multicolumn{5}{c}{ Aroma }
& \multicolumn{5}{c}{ Palate } \\
\cmidrule(lr){2-6}
\cmidrule(lr){7-11}
 &Acc $(\uparrow)$  & P $(\uparrow)$& R $(\uparrow)$ & F1 $(\uparrow)$ & FDR $(\downarrow)$& Acc $(\uparrow)$ & P $(\uparrow)$ & R $(\uparrow)$& F1 $(\uparrow)$ &FDR $(\downarrow)$\\
\midrule
VIB  & $80.2(1.8)$ &$30.0(4.3)$ &$17.2(3.1)$ & $20.1(3.5)$ & $97.6(5.3)$ & $75.2(3.8)$ &$26.8(4.5)$ & $20.7( 4.0)$ & $21.6(3.8)$ & $23.2(1.6)$ \\
RNP & $79.0(1.0)$ &$44.1(5.3)$ &$12.3(1.0)$ & $18.4(1.5)$& $97.4(3.0)$ & $72.0(2.2)$ &$22.8(2.6)$ &$18.1(1.2)$ & $18.6(1.6)$& $55.7(1.9)$ \\
FR   & $84.5(1.0)$ &$38.7(3.7)$ &$22.7(2.2)$ & $27.0(2.7)$ & $41.2(2.2)$& $76.2(3.5)$ &$25.4(3.7)$ & $22.0(4.0)$& $21.7(4.3)$ &$11.6(3.1)$ \\
A2R & $82.0(1.5)$ &$44.6(3.4)$ &$20.8(1.7)$& $26.8(3.2)$&$60.7(5.9)$& $73.5(3.8)$&$28.9(1.2)$&$21.8(3.2)$ & $22.0(3.4)$ &$39.7(3.3)$\\
INVRAT & $84.0(1.5)$ & $42.0(2.6)$ &$22.9(1.8)$ & $27.9(2.4)$ &$40.3(3.1)$& $\mathbf{76.5(3.6)}$ &$28.2(2.4)$&$22.5(2.7)$& $22.5(3.0)$ &$7.2(1.7)$\\
CR(ours)  & $\mathbf{85.0(2.1)}$ &$\mathbf{55.3( 3.1)}$  &$\mathbf{31.5(1.8)}$& $\mathbf{37.8(2.1)}$&$\mathbf{37.9(0.7)}$ &  $73.0(2.3)$ & $\mathbf{29.4(3.5)}$ & $\mathbf{22.6(3.8)}$& $\mathbf{23.8 (3.3)}$ & $\mathbf{3.0(1.1)}$ \\
\bottomrule
\end{tabular}
}
\label{beer noise}
\end{table*}
 

\subsection{Real Data Experiments}

\label{real data exp}
We evaluate our method on three real datasets, Beer review, Hotel review, and GA data. For Beer and Hotel review data, based on summary in \cref{data} \cref{beer stat summary}, we select Top-$10\%$ tokens in the test stage and the way of choosing $k$ during evaluation is in Appendix \ref{select k}. It is shown in Table \ref{beer real} that our method achieves a consistently better performance than baselines in most metrics for Beer review data. Results for hotel review data are in Appendix \ref{hotel-result}. Specifically, our method demonstrates a significant improvement over VIB which is an ablation study and indicates our causal component contribution to the superior performance.

\begin{remark}
\label{update fr}
\textbf{New Results for FR}

\color{black}For the FR method, we found that when FR used the same low learning rate as CR, its prediction accuracy decreased to some extent. Therefore, in the original experiment, we used a relatively higher learning rate for FR. However, we discovered that when using a lower learning rate or fewer training epochs, FR can achieve relatively good interpretability results. Therefore, we have included the experimental results of FR using other hyperparameters in the appendix \ref{more fr} to make experiments more comprehensive.
\end{remark}

We calculate empirical $\underline{\operatorname{CPNS}}$ on the test dataset as shown in \cref{data cpns}. We find that our approach always obtains the highest values in three aspects that match our expectation because one goal of our objective function is to maximize the $\underline{\operatorname{CPNS}}$. This explains why our method has superior performance and indicates that $\underline{\operatorname{CPNS}}$ is effective to find true rationales under the in-distribution setting. Examples of generated rationales are shown in \cref{one vis beer examples} and more examples are provided in \cref{all Visualization beer Results}. We also conduct sensitivity analyses for hyperparameters of causality constraints in \cref{sensitivity} and the results show the performance is insensitive to $k$ and $\mu$ when $\mu$ is not too large.  

For the GA dataset, as shown in \cref{ga real}, our method is slightly better than baselines in terms of prediction performance. We further examine the generated rationales on GA patients and observe that our causal rationalization could provide better clinically meaningful explanations, with visualized examples in \cref{all Visualization ga Results}. This shows that CR can provide more trustworthy explanations for EHR data.

\subsection{Synthetic Experiments}
\label{synthetic}

\noindent \textbf{Beer-Spurious}. We include spurious correlation into the Beer dataset by randomly appending spurious punctuation. We follow a similar setup in \citet{chang2020invariant} and \citet{yu2021understanding} to append punctuation “,” and “.” at the beginning of the first sentence with the following distributions:
\resizebox{.99\linewidth}{!}{
\begin{minipage}{\linewidth}
\vspace{-0.4cm}
\begin{align*}
&\mathrm P(\text { append ","} \mid Y=1)=\mathrm P(\text { append "."} \mid Y=0)=\alpha_1, \\
&\mathrm P(\text { append  "."} \mid Y=1)=\mathrm P(\text { append ","} \mid Y=0)=1-\alpha_1.
\end{align*} 
\end{minipage}
}

Here we set $\alpha_1=0.8$. Intuitively, since the first sentence contains the appended punctuation with a strong spurious correlation, we expect the association-based rationalization approach to capture such a clue and our causality-based method can avoid selecting spurious tokens. Since for many review comments, the first sentence is usually about the appearance aspect, here we only utilize aroma and palate aspects as \citet{yu2021understanding}. We inject tokens into the training and validation set, then we can evaluate OOD performance on the unchanged test set. See more details in Appendix \ref{beer_spurious}.

\begin{table}[!thp] 
\centering
\caption{Results on the GA data. Top-$5\%$ tokens are highlighted for evaluation.}
\vskip -0.1in
    \resizebox{0.98\columnwidth}{!}{
\begin{tabular}{lcccc} 
\toprule
&CR (ours) & RNP&VIB&FR  \\
\midrule
AUC $(\uparrow)$& $84.3 (\pm0.2)$ & $83.3(\pm 0.8)$ & $84.00(\pm 0.5)$ & $84.2(\pm1.0)$   \\
\bottomrule
\end{tabular}
}
\label{ga real}
\end{table}

\begin{figure}[!t]
        \centering
        \includegraphics[width=0.5\textwidth]{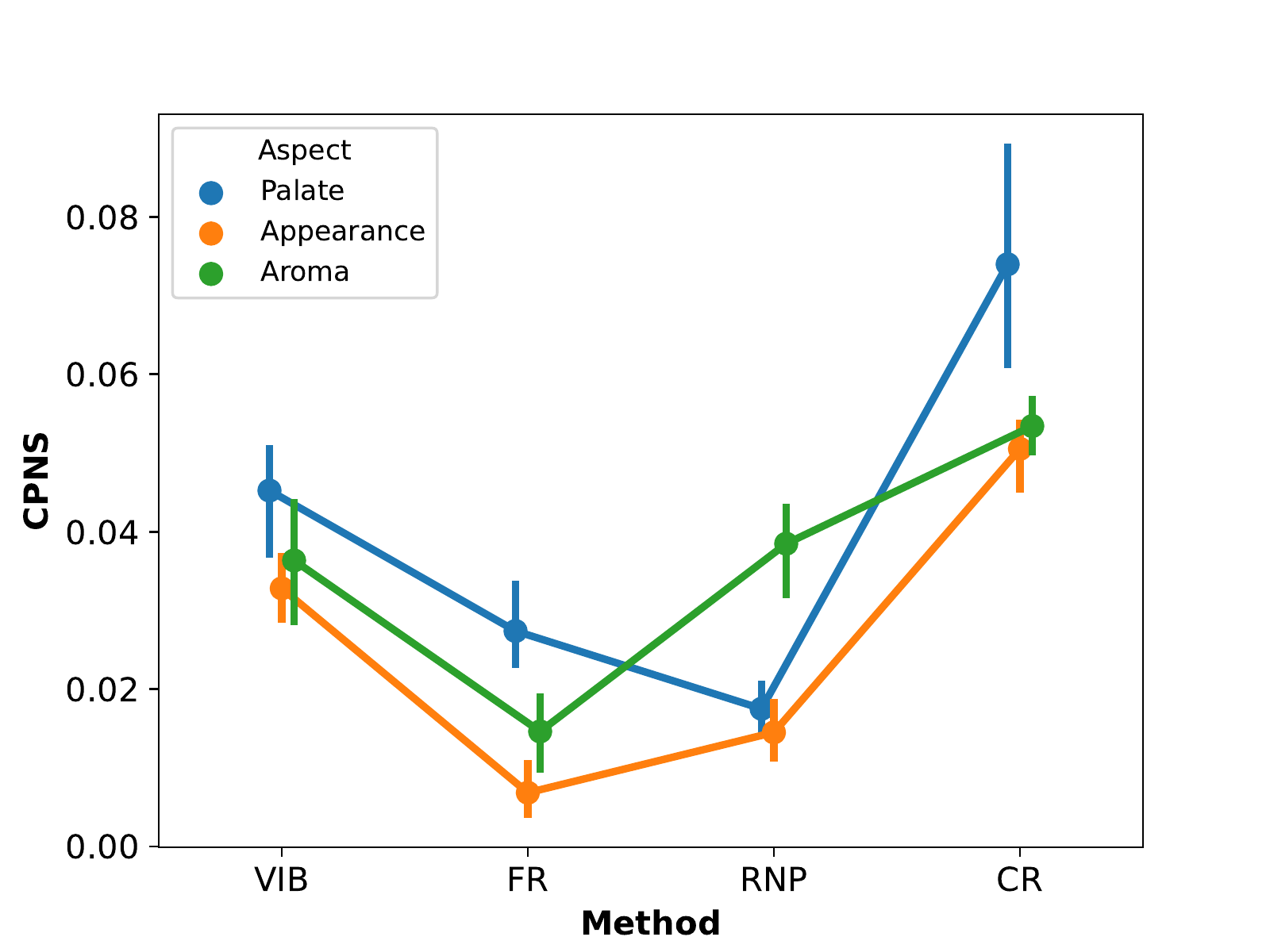}
        \caption{Estimated lower bound of CPNS of ours and baseline methods on the test set of the Beer review data.}
        \label{data cpns}
\end{figure} 

\begin{table*}[!htbp]
\centering
\caption{Examples of generated rationales. Human annotations are \blackulns{underlined} and rationales obtained from ours and baseline methods are highlighted by different colors. We find that the rationales of our method better align with human annotations.}
\vskip -0.1in
    \resizebox{0.9\textwidth}{!}{
    \begin{minipage}{1.2\textwidth}
            
    \begin{tabular}{ p{4.8cm}  p{4.8cm}  p{4.8cm}  p{4.8cm}}
    
        \toprule
\textbf{CR}      
& \textbf{VIB}  
& \textbf{RNP}
& \textbf{FR}
\\\midrule
\textbf{Aspect}: Appearance
\newline
\textbf{Label}: Positive
\textbf{Pred}: Positive
\newline
poured from bottle into shaker pint glass . \redulns{a :} \redulns{pale tan-yellow} \blackulns{with} \redulns{very thin white head} \blackulns{that} \redulns{quickly disappears} . s : malt , caramel ... and skunk t : very bad . hint of malt . mostly just tastes bad though . m : acidic , thin , and watery d : not drinkable . unbelievably bad . stay away ... you have several thousand other beers that taste better to spend you money on . do n't be a fool like me . 
& 
\textbf{Aspect}: Appearance 
\newline
\textbf{Label}: Positive
\textbf{Pred}: Positive
\newline
poured from bottle into shaker pint \textcolor{blue}{glass} . \blackulns{a :}
\blueulns{pale tan-yellow} \blackulns{with very} \blueulns{thin white} \blackulns{head that} \blueulns{quickly disappears} . s : malt , caramel ... and skunk t : very \textcolor{blue}{bad} . hint of malt . mostly just tastes bad though . m : acidic , \textcolor{blue}{thin} , and \textcolor{blue}{watery} d : not drinkable . unbelievably bad . stay away ... you have several thousand other beers that taste better to spend you money on . do n't be a fool like me .      
& 
\textbf{Aspect}: Appearance 
\newline
\textbf{Label}: Positive
\textbf{Pred}: Positive
\newline
poured from bottle into shaker pint glass . \blackulns{a :} \orangeulns{pale tan-yellow} \blackulns{with very} \orangeulns{thin} \blackulns{white head that quickly} \orangeulns{disappears} . s : malt , caramel ... and skunk t : very bad . hint of malt . mostly just tastes bad though . m : acidic , thin , and \textcolor{orange}{watery} d : \textcolor{orange}{not} drinkable . unbelievably bad . stay away ... you have several thousand other \textcolor{orange}{beers} that taste better to spend you money on . do n't be a fool like me .     
&
\textbf{Aspect}: Appearance 
\newline
\textbf{Label}: Positive
\textbf{Pred}: Positive
\newline
poured from bottle into shaker pint \textcolor{green}{glass} . \blackulns{a :} \greenulns{pale tan-yellow} \blackulns{with very} \greenulns{thin} \blackulns{white head that quickly} \greenulns{disappears} \textcolor{green}{.} s : malt , caramel ... and skunk t : very bad . hint of malt . mostly just tastes \textcolor{green}{bad} though . m : acidic , thin , and \textcolor{green}{watery} d : \textcolor{green}{not} drinkable . unbelievably \textcolor{green}{bad} . stay away ... you have several thousand other beers that taste better to spend you money on . do n't be a fool like me .  \\
\bottomrule
 \label{one vis beer examples}
    \end{tabular}
    \end{minipage}
    }
    \vskip -0.25in
\end{table*}

\begin{figure}[!htbp]
\centering
        \centering
            \vskip -0.2in
            \includegraphics[width=\columnwidth]{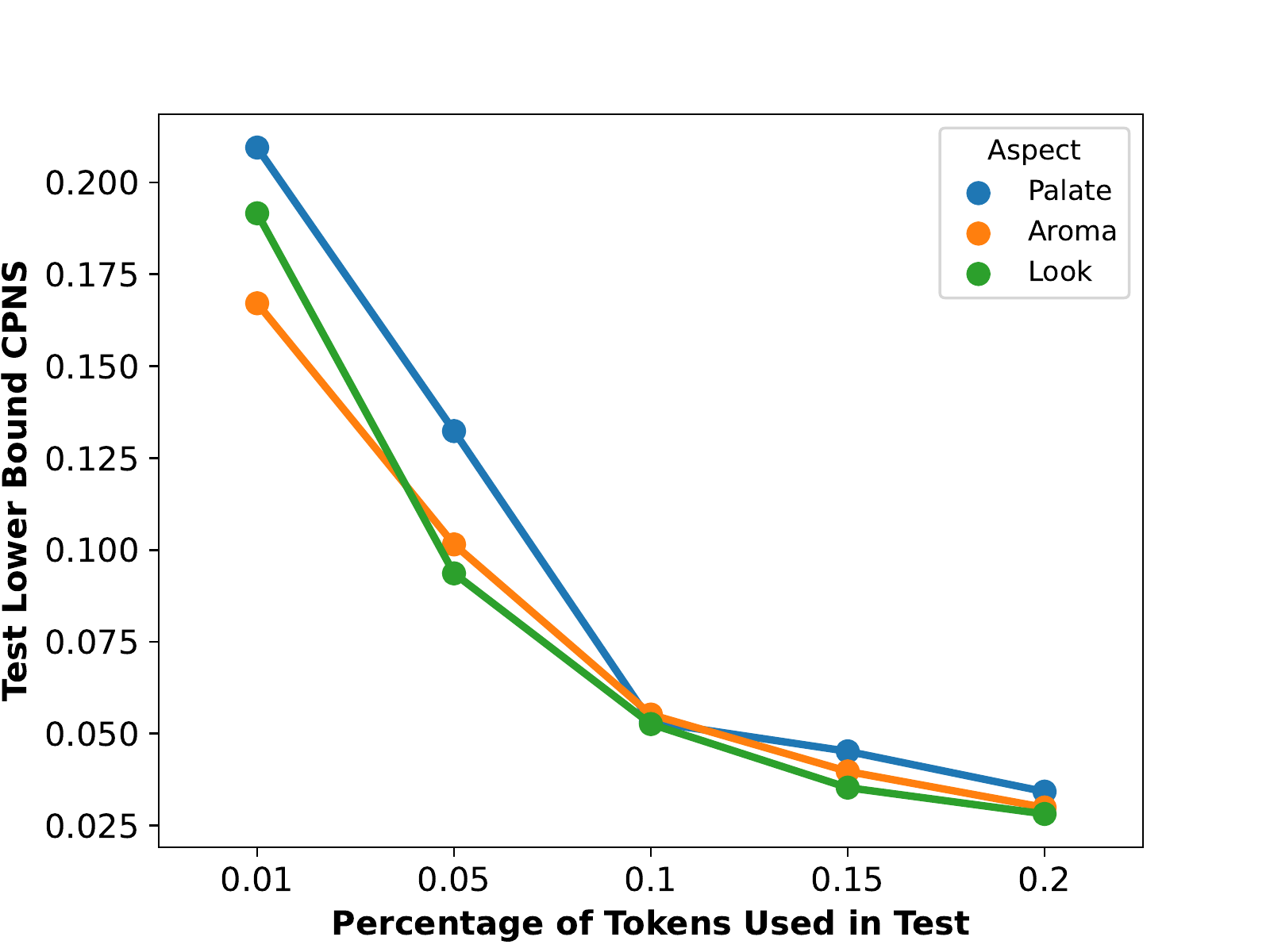}
            \caption{The lower bound of $\operatorname{CPNS}$ with different highlighted lengths during test.}
            \vskip -0.1in
                    \label{hilight-length}
\end{figure}

From Table \ref{beer noise}, our causal rationalization outperforms baseline methods in most aspects and metrics. Especially, our method selects only $37.9\%$ and $3.0\%$ injected tokens on the validation set for two aspects. RNP and VIB don't align well with human annotations and have low prediction accuracy because they always select spurious tokens. We notice it's harder to avoid selecting spurious tokens for the aroma aspect than the palate. Our method is more robust when handling spurious correlation and shows better generalization performance, which indicates $\underline{\operatorname{CPNS}}$ can help identify true rationales under an out-of-distribution setting. 

We evaluate out-of-distribution performance on the unchanged testing set, so the results in Table \ref{beer noise} do suggest that FR is better at avoiding the selection of spurious tokens than VIB and RNP under noise-injected scenarios and has better generalization. The results that VIB is better than FR in Table \ref{beer real} don’t conflict with the previous finding because Table \ref{beer real} evaluates in-distribution learning ability and VIB can be regarded as more accurately extracting rationales without distribution shift. Figure 6 presents an estimated lower bound of the $\operatorname{CPNS}$ for our method and the baseline methods on the test set of the Beer review data. While it is true that FR has the lowest $\operatorname{CPNS}$ in Figure \ref{data cpns} this does not negate the importance of $\operatorname{CPNS}$. It is crucial to note that the $\operatorname{CPNS}$ value estimates how well a model extracts necessary and sufficient rationales, but it is not the sole indicator of a model's overall performance. The primary goal of FR is to avoid selecting spurious tokens, which is evidenced by its lower FDR in Table \ref{beer noise}, demonstrating its better generalization under noise-injected scenarios. We can argue that the discrepancy between $\operatorname{CPNS}$ and FDR for FR may arise due to the fact that FR is more conservative in selecting tokens as rationales. This may lead to a lower $\operatorname{CPNS}$, as FR may miss some necessary tokens, but at the same time, it avoids selecting spurious tokens, thus resulting in a lower FDR.

 \subsection{How Highlighted Length Influence CPNS}
\label{test ratio}

We want to know how highlighted length during evaluation can influence $\underline{\operatorname{CPNS}}$. We evaluate $\underline{\operatorname{CPNS}}$ for three aspects with different percentages of tokens from $\{1\%,5\%,10\%,15\%,20\%\}$ as shown in \cref{hilight-length}. It can be seen that as the highlighted length increases, the estimated value decreases, which matches our expectations. For $\underline{\operatorname{CPNS}}$, it consists of two types of conditional probabilities $P(Y=y \mid Z_{j}=z_{j},\boldsymbol{Z}_{-j}=\boldsymbol{z}_{-j},\boldsymbol{X}= \boldsymbol{x})$ and $-P(Y=y \mid Z_{j} \neq z_{j},\boldsymbol{Z}_{-j}=\boldsymbol{z}_{-j},\boldsymbol{X}= \boldsymbol{x})$. If more events are given, namely the dimension of $\boldsymbol{Z}_{-j}$ increases, a flip of a single selection $Z_{j}$ would bring less information, and hence the difference between two probabilities would decrease. The results indicate that we should always describe and compare $\underline{\operatorname{CPNS}}$ of rationalization approaches using the same highlighted length.

\section{Conclusion}
This work proposes a novel rationalization approach to find causal interpretations for sentiment analysis tasks and clinical information extraction. We formally define the non-spuriousness and efficiency of rationales from a causal inference perspective and propose a practically useful algorithm. Moreover, we show the superior performance of our causality-based rationalization compared to state-of-the-art methods. The main limitation of our method is that $\operatorname{CPNS}$ is defined on the token-level and the computational cost is high when there are many tokens hence our method is not scalable well to long-text data. In future work, an interesting direction would be to define $\operatorname{CPNS}$ on the clause/sentence-level rationales. This would not only make the computation more feasible but also extracts higher-level units of meaning which improve the interpretability of the model's decisions.

\bibliography{ref}
\bibliographystyle{icml2023}


\newpage
\appendix


\onecolumn

\section{More Details of Probability of Causation for Rationales}

\subsection{More Details of Definitions \ref{ps}-\ref{pns}}
\label{clarification}
The probability of sufficiency (PS) for a binary label $Y$ and a binary feature $Z$ is defined as $P\{Y(Z=True)=True|Y=False, Z=False)$, which means given the fact that we observed the false label $Y$ with a false feature $Z$, what is the probability of the label turning to be true if we have had the chance to set the feature to be true. This probability thus describes the sufficiency of the feature $Z$ to be true to obtain a true label. Now, moving to our Definition \ref{ps} proposed for the rationales, it means the probability of $Y=y$ when changing the selection to be $Z=z$ given text $X=x$ with the already observed selection $Z\neq z$ and the label $Y \neq y$. In other words, PS gives the probability of setting $z$ would produce $y$ in a situation where $z$ and $y$ are in fact absent given $x$. It describes the capacity of the rationales to “produce” the label. On the other hand, the probability of necessity (PN) for a binary label and a binary feature is $P\{Y(Z=False)=False|Y=True, Z=True)$, which means given the fact that we observed the true label $Y$ with a true feature $Z$, what is the probability of the label turning to be false if we have had the chance to set the feature to be false. This probability thus describes the necessity of the feature $Z$ to be true to obtain a true label. Following a similar logic, Definition \ref{pn} generalizes the PN score in the bivariate setting for rationales to describe the selection $Z$ being a necessary cause of the label. Combining PN and PS yields the probability of necessity and sufficiency (PNS) for rationales in Definition \ref{pns} to comprehensively characterize the importance of rationales in causally determining the label. Those defined counterfactual quantities provide a formal way of measuring whether one event A is the necessary or sufficient cause of another event B. 

\subsection{Significance of Definitions \ref{ps}-\ref{df_single_cpns} and Why PNS Means Causality}
\label{significance}

The significance of all the proposed definitions of the probability of causation (POC) is three-fold. First, we generalized the classical definitions in \citet{pearl2000causality} (where they only consider one binary outcome and one binary feature) for rationales to allow a selection of words as the feature input. Secondly, we align POC with the newly proposed structural causal model for rationalization, with additional conditioning on the texts. Our definitions imply that the input texts are fixed and thus POC is mainly detected through interventions on selected rationales. Lastly, to accommodate the task of rationalization, we further define $CPNS_j$ for the $j$-th selected token in Definition \ref{df_single_cpns}, which allows us to simultaneously quantify the sufficiency and necessity of an individual rationale.

We also illustrate why PNS Means Causality. First of all, we regard the underlying labeling process as a structure causal model shown in Figure 4, and thus the true selection $Z$ can be seen as the cause of the label. Changing the underlying true $Z$ selection to the opposite should also change the value of the label accordingly. This is aligned with the definition of PNS and thus it represents the causality of how the selected rationales determine the label. If a selection of tokens is necessary and sufficient causes the label, it should have a high PNS to reflect its high necessity and sufficiency in determining the label. We expect our rationalization approach can capture the underlying causal selection, so that’s why we focus on optimizing PNS and CPNS.

\subsection{Discussion of Geometric Average in Definition \ref{df_ocpns}}
\label{geometric}
We use the geometric mean because we want CPNS over selected rationales in Definition \ref{df_ocpns} to be a likelihood, which is a product of all individual rationales’ $\operatorname{CPNS}$ scores in the selection. Yet, such a simple product is not ideal owing to the heterogeneous length of the texts. As the length increases, the number of selected rationales increases (because it’s $k\%$ of raw text length), and the product would be lower by noting the probability ranging from 0 to 1. This is undesirable because we don’t want to text length to be an influencing factor. Hence, we normalize the likelihood, leading to the geometric mean as presented in the current paper.

\subsection{Clarification on Token-level $\operatorname{CPNS}$ and Reasoning}
\label{token-level}
 We propose Definition \ref{df_single_cpns} to assess the causality of each token, which can be estimated by comparing the conditional probability of the label with and without this token (two counterfactual realities), as stated in Theorem \ref{theorem}. Therefore, to assess the causality of the rationale (i.e., a selection of tokens), the most natural way is to define the $\operatorname{CPNS}$ directly for this rationale by comparing the conditional probability of the label with the current selection versus that with counterfactual selections. However, for a selection consisting of $r$ tokens, the counterfactual realities yield $2^r-1$, which leads to a computational challenge when $r$ is large. To overcome this issue, we alternatively propose Definition \ref{df_ocpns} to combine all individual-level $\operatorname{CPNS}$ scores in the selection to reflect the causality of the rationale, which instead yields a polynomial computational complexity of $\mathcal{O}(r)$. Suppose the spillover effects (i.e., the words have dependence and the effect of the joint of words cannot simply be written as the sum of effects of single words) are uniform across words and sentences, with the Geometric Average in Definition \ref{df_ocpns}, then we have the combined $\operatorname{CPNS}$ to approximate the $\operatorname{CPNS}$ of this rationale. Thus, this score helps to determine the causality of the rationale. When there is no spillover effect, Definition \ref{df_single_cpns} can be viewed as the $\operatorname{CPNS}$ of a rationale directly. Notably, we \textbf{do not calculate} $\operatorname{CPNS}_j$ for all tokens. Rather, we focus on a selection of tokens indexed by $r$, where the index set $r$ can be determined by the selector $g_{\theta}$. The demonstration of the toy example mainly explains why $\operatorname{CPNS}$ is useful in identifying true causes in the presence of spurious features.

 \section{Structural Causal Model for Rationalization under Potential Dependences}
\label{cla-scm}
We firstly clarify that according to Figure \ref{causal diagram} and the proposed structural causal model for rationalization, the entire rationale as a selection of true important tokens $\mathbf{Z}$ is determined by texts $\mathbf{X}$, however, we do allow exogenous variables $N_Z$ to model possible dependence among the elements of $\mathbf{Z}$ (i.e., $z_i$). We understand that in practice, the rationale selection process may involve sequential labeling, where the selection of one token can influence the subsequent selections.

\noindent \textbf{Impact of Potential Dependence on Our Method.}
To address the concern of whether the potential dependence or sequential labeling would affect our method, we argue that our method is still valid, but with conditions. Specifically, recall that we propose Definition \ref{df_single_cpns} to assess the causality of each token, which can be estimated by comparing the conditional probability of the label with and without this token (two counterfactual realities), as stated in Theorem \ref{theorem}. Therefore, to assess the causality of the rationale (i.e., a selection of tokens), the most natural way is to define the $\operatorname{CPNS}$ directly for this rationale by comparing the conditional probability of the label with the current selection versus that with counterfactual selections. Such a definition would allow possible dependence among the elements of $\mathbf{Z}$ (i.e., $z_i$).

However, for a selection consisting of $r$ tokens, the counterfactual realities yield $2^r-1$, which leads to a computational challenge when $r$ is large. To overcome this issue, we alternatively propose Definition \ref{df_ocpns} to combine all individual-level $\operatorname{CPNS}$ scores in the selection to reflect the causality of the rationale, which instead yields a polynomial computational complexity of $\mathcal{O}(r)$. The proposed alternative approach is valid under potential dependence including the following cases: 1. Suppose the spillover effects (i.e., the words have dependence and the effect of the joint of words cannot simply be written as the sum of effects of single words) are uniform across words and sentences, with the Geometric Average in Definition \ref{df_ocpns}, then we have the combined $\operatorname{CPNS}$ to approximate the $\operatorname{CPNS}$ of this rationale. Thus, this score helps to determine the causality of the rationale. 2. When there is no spillover effect, Definition \ref{df_ocpns} can be viewed as the $\operatorname{CPNS}$ of a rationale directly. 3. In addition, suppose there exists conditional independence among words (considered in \citet{joshi2022all}), our proposed combined $\operatorname{CPNS}$ is equivalent to the $\operatorname{CPNS}$ of a rationale as well.

In conclusion, our method can still be applicable under the presence of potential dependence or sequential labeling, with certain conditions on the spillover effects of the dependency. Our approach to assessing the causality of rationales using the combined CPNS allows us to account for possible dependencies among the elements of $\mathbf{Z}$ while maintaining computational efficiency.

\section{Data Prepossessing and Summary }

\label{data}

\noindent \textbf{Beer Review Data.}  We use the publicly available version of the Beer review dataset also adopted by \citet{bao2018deriving} and \citet{chen2022can}. This dataset is cleaned by the previous authors and is a subset of the raw BeerAdvocate review dataset \cite{mcauley2012learning}. Following the same evaluation protocol of some previous works \citep[see e.g.,][]{bao2018deriving,yu2019rethinking,chang2020invariant,chen2022can}, we convert their original scores which are in the scale of $[0,1]$ into binary labels. Specifically, reviews with ratings $\leq 0.4$ are labeled as negative and those with $\geq 0.6$ are labeled as positive. We follow the same train/validation/test split as \citet{chen2022can} and it is summarized in Table \ref{beer number summary}. To make computation more feasible, except for the raw dataset, we create a short-text version of the dataset by filtering the texts over a length of $120$. Table \ref{beer stat summary} summarizes the statistics of the Beer review dataset. 

\begin{table}[!htbp]
\caption{The split of the dataset.}
\centering
\begin{tabular}{cccc}
\toprule
Short & Train & Val & Test\\
\midrule
Beer (Appearance) & $15932$ & $3757$ & $200$ \\
 Beer (Aroma) & $14085$ & $2928$& $200$ \\
 Beer (Palate) & $9592$ & $2294$ & $200$  \\
\bottomrule
\end{tabular}
\label{beer number summary}
\end{table}

\begin{table}[!htbp]
\caption{Dataset details, with rationale length ratios included for datasets where they are available.}
\centering
\begin{tabular}{lccc}
\toprule Short & Len & Rationale(\%) & N \\
\midrule Beer (Appearance) & $88.94$ & $19.2 $ & 19889 \\
 Beer (Aroma) & $89.92$ & $15.9$& 17033 \\
 Beer (Palate) & $90.72$ & $12.7$ & 12086  \\
\bottomrule
\end{tabular}
\label{beer stat summary}
\end{table}

\noindent \textbf{Hotel Review Data.} The Hotel review data we used were first proposed by \citet{wang2010latent} and we adopted processed one from \citet{bao2018deriving}.  Table \ref{hotel stat summary} summarizes the statistics of the location aspect of the Hotel review dataset.

\begin{table}[!htbp]
\caption{Dataset details for Hotel review data, with rationale length ratios included for datasets where they are available.}
\centering
\begin{tabular}{lccccc}
\toprule Aspect  & Train & Val & Test& Len & Rationale(\%)  \\
\midrule Location & $14472$ &$1883$ & $200$  &$708.7$ & $10.3$ \\
\bottomrule
\end{tabular}
\label{hotel stat summary}
\end{table}

\noindent \textbf{GA Data.} 
The proprietary GA dataset used in this study includes the medical claim records (diagnosis, prescriptions, and procedures) of 329,023 patients who were diagnosed as GA from 2018 to 2021 in the US, as well as those of additional 991,946 patients who have at least one of GA risk factors. Since patients have long medical sequences over years, here we only extract their most recent two years' visits. Then we further select patients with sequence lengths between 100 and 150. Finally, we sample 10000 positive patients and 30000 negative patients from the cohort to construct our dataset and \cref{GA summary} illustrates the division of the data into training, validation, and test sets.

\begin{table}[!htbp]
\caption{The split of the dataset.}
\centering
\begin{tabular}{cccccc}
\toprule Short & Train & Val & Test & Total & Len\\
\midrule GA & $20000$ & $10000$ & $10000$ & $40000$ &122.31\\
\bottomrule
\end{tabular}
\label{GA summary}
\end{table}




 
\section{Implementation Details}\label{implt_appen}
For the Beer review data, we use two BERT-base-uncased as the selector and the predictor components for rationalization approaches. Those modules are initialized with pre-trained Bert. 

A few challenges rise prior to directly including INVRAT for a fair comparison. Firstly, as discussed in the related work, INVRAT relies on multiple environments while our method and other baselines focus on a single environment. Second, their method highly depends on the construction of such multiple environments, yet, there is no principled guidance on how to select environments for INVRAT. Third, \citet{chang2020invariant} trained a simple linear regression model to predict the rating of the target aspect given the ratings of all the other aspects to generate the environments. This dataset is not publicly available. In contrast, for the public Beer dataset we used, a comment for one aspect doesn’t have scores for other aspects, which means we can’t simply utilize the experimental design of \citet{chang2020invariant}  to select suitable environments. To address those issues, firstly we impute the missing scores based on aspect-specific prediction models which are trained on text data with their single provided aspect and then we follow the same way as \citet{chang2020invariant} to select environments.

For the GA data, we use the same architecture and the only difference is we replace the word embedding matrix with a randomly initialized health diagnosis code embedding and the embedding is trained jointly with other modules. For all experiments, we utilize a batch size of $256$ and choose the learning rate $\alpha \in \{1e-5,5e-4,1e-4\}$. We train for $10$ epochs all the datasets. For training the causal component, we tune the values of the Lagrangian multiplier $\mu\in\{0.01,0.1,1\}$ and set $k=5$. We set the temperature of Gumbel-softmax to be $0.5$. For our final evaluation, we highlight Top-$10\%$ tokens as the rationales for the Beer review data and Top-$5\%$ for the GA data. We conduct our experiments over $5$ random seeds and calculate the mean and standard deviation of metrics. All of our experiments are conducted with PyTorch on 4 V100 GPU. Our code is publicly available online.\footnote{\href{https://github.com/onepounchman/Causal-Retionalization}{https://github.com/onepounchman/Causal-Retionalization}.}

\section{How to Select $k\%$ during Evaluation}
\label{select k}
\noindent \textbf{Rationale for choosing $k=10\%$ for the Beer review dataset:} As shown in Appendix \ref{data} Table \ref{beer stat summary}, the true rationales in the Beer review dataset are between $10\%$ and $20\%$ of the total tokens. Therefore, we choose $k=10\%$ as a reasonable threshold to represent the ground truth for this dataset, ensuring that we capture a significant portion of the true important tokens in our evaluation.

\noindent \textbf{Rationale for choosing $k=5\%$ for the GA dataset:} The GA dataset consists of medical claim data, and a substantial portion of records (around $10\%$) are related to administrative and billing purposes, for example, codes for office visits or inpatient/outpatient admin records. These records offer limited insights into patients' disease progression and are less relevant as rationales. Given the smaller pool of meaningful rationales in the GA dataset compared to the Beer review dataset, we set a lower threshold ratio ($k=5\%$) for this dataset.

In conclusion, the choice of different top token ratios for the Beer review dataset and GA dataset in Experiment \ref{real data exp} is based on the characteristics and ground truth of each dataset. Our aim is to ensure a fair evaluation of the models' performance in extracting meaningful rationales from the texts while taking into account the specific context and content of each dataset.

\section{More Elaboration on Spurious Experiments}\label{beer_spurious}
\noindent \textbf{Punctuations as An OOD Scenario.} While it is true that adding punctuation tokens does not change the meaning of the sentence, the distribution of spurious tokens is changed. The objective of the experiment in Section \ref{synthetic} is to evaluate the model's generalization under different conditions. By injecting punctuation tokens based on the label (as described in Section \ref{synthetic}), we introduce spurious correlations that the model may exploit during training. These spurious tokens can be regarded as short-cuts that can potentially mislead rationalization methods. With and without punctuation are two scenarios representing whether short-cuts exit or not, we consider this as an OOD setting.

\noindent \textbf{Clarification on Training/Validation/Test Data.} In this experiment, we add punctuation tokens only to the training and validation data, keeping the test data unchanged. This setup allows us to examine the models' ability to generalize in an OOD setting, where the distribution of spurious tokens in the training and validation data is different from that in the test data. By keeping the test data free of injected punctuations, we can evaluate how well the models perform when faced with a scenario where the short-cuts present in the training and validation data are absent in the test data.

\section{Experimental Results}

\subsection{Results Analyses}{
In conclusion, the experimental results demonstrate that FR is indeed better at avoiding the selection of spurious tokens and has better generalization under noise-injected scenarios. The differences between the findings in Tables 1, Table 2, and Figure 6 highlight the distinct evaluation scenarios (in-distribution vs. out-of-distribution) and emphasize the importance of considering multiple performance metrics (F1, FDR, and CPNS) to obtain a comprehensive understanding of the models' behavior.

}

\subsection{More Extensive Results for FR}
\label{more fr}

More experiments for FR are conducted by adjusting the learning rate and training epochs. The results are in Tables \ref{new fr1} and \ref{new fr2}. 

In Table \ref{new fr1}, it is evident that adjusting the learning rate of FR to either 5e-6 or 1e-5 significantly improves the F1 scores of FR. Although the accuracy somewhat decreased, this trade-off is acceptable considering the notable improvement in F1-score (e.g., $-2.4\%\rightarrow 8.4\%$).

In Table \ref{new fr2}, FR converges much faster than CR, so an additional setting by halving the training epochs is considered here. There is a substantial improvement in the F1 score across all three settings. However, there is also a significant drop in accuracy for FR. 

These additional results can help readers gain a clearer understanding of the effectiveness of both FR and CR. 

\begin{table}
\centering
\caption{Results of experiments corresponding to Table \ref{beer real}. "*": raw results. "()": the standard deviation. "improvement": $\frac{F R^A-F R^*}{F R^*}$. "lr": learning rate. “e”: training epochs. $\mathrm{FR}^A$ and $\mathrm{FR}^B$ are FR method with different hypeparamters. The worst $\mathrm{F} 1$-scores are highlighted in \textbf{bold}.}
\begin{tabular}{c|cc|cc|cc}
\hline \multirow{2}{*}{ Methods } & \multicolumn{2}{c|}{ Appearance } & \multicolumn{2}{c|}{ Aroma } & \multicolumn{2}{c}{ Palate } \\
\cline { 2 - 7 } & Acc & F1 & Acc & F1 & Acc & F1 \\
\hline FR $^*(\mathrm{lr}=5 \mathrm{e}-5)$ & $94.5(0.6)$ &$\boldsymbol{23.7(1.1)}$ & $86.0(1.6)$ & $\boldsymbol{28.0(1.9)}$ & $77.0(1.9)$ & \boldsymbol{$22.4(1.6)$} \\
\hline $\mathrm{FR}^A(\mathrm{lr}=5 \mathrm{e}-6)$ & $92.2(2.1)$ & $25.7(1.0)$ & $78.1(5.1)$ & $34.5(1.2)$ & $72.5(1.9)$ & $27.2(1.9)$ \\
improvement & $-2.4 \%$ & $8.4 \%$ & $-9.2 \%$ & $23.2 \%$ & $-5.8 \%$ & $21.3 \%$ \\
\hline $\mathrm{FR}^B(1 \mathrm{r}=1 \mathrm{e}-5)$ & $90(4.8)$ & $26.3(2.3)$ & $81.7(6.4)$ & $35.4(0.9)$ & $80(1.3)$ & $31.8(2.7)$ \\
improvement & $-4.8 \%$ & $11.0 \%$ & $-5.0 \%$ & $26.6 \%$ & $3.9 \%$ & $42.1 \%$ \\
\hline
\end{tabular}
\label{new fr1}
\end{table}

\begin{table}
\centering
\caption{Results of experiments corresponding to Table \ref{beer noise}. "*": raw results. "()": the standard deviation. "improvement": $\frac{F R^A-F R^*}{F R^*}$. "lr": learning rate. “e”: training epochs. $\mathrm{FR}^A$ and $\mathrm{FR}^B$ are FR method with different hypeparamters. The worst $\mathrm{F} 1$-scores are highlighted in \textbf{bold}.}
\begin{tabular}{c|cc|cc}
\hline \multirow{2}{*}{ Methods } & \multicolumn{2}{c|}{ Aroma } & \multicolumn{2}{c}{ Palate } \\
\cline { 2 - 5 } & Acc & $\mathrm{F} 1$ & Acc & $\mathrm{F} 1$ \\
\hline $\mathrm{FR}^*(\mathrm{lr}=5 \mathrm{e}-5, \mathrm{e}=10)$ & $84.5(1.0)$ & \boldsymbol{$27.0(2.7)$} & $76.2(3.5)$ & \boldsymbol{$21.7(4.3)$} \\
\hline $\mathrm{FR}^A(\mathrm{lr}=5 \mathrm{e}-6, \mathrm{e}=10)$ & $73.2(4.2)$ & $40.2(1.0)$ & $71.8(4.0)$ & $29.3(2.0)$ \\
improvement & $-13.3 \%$ & $48.9 \%$ & $-5.8 \%$ & $35.2 \%$ \\
\hline $\mathrm{FR}^B(\mathrm{lr}=1 \mathrm{e}-5, \mathrm{e}=10)$ & $65.3(2.9)$ & $42.7(0.9)$ & $71.7(5.2)$ & $34.7(0.8)$ \\
improvement & $-22.7 \%$ & $58.2 \%$ & $-5.9 \%$ & $59.9 \%$ \\
\hline $\mathrm{FR}^C(\mathrm{lr}=5 \mathrm{e}-5, \mathrm{e}=5)$ & $78.0(9.1)$ & $39.0(1.0)$ & $76.3(3.0)$ & $32.6(1.9)$ \\
improvement & $-7.7 \%$ & $44.3 \%$ & $0.1 \%$ & $50.1 \%$ \\
\hline
\end{tabular}
\label{new fr2}
\end{table}

\subsection{Beer Review Results after Adding Continuous Constarint}
\label{add-cont}
From Table \ref{beer conti}, we observe that our method, when applied with the continuity constraint, continues to perform well, suggesting that the continuity constraint does not negatively impact our method's effectiveness.

\begin{table*}[t] 
\centering
\caption{Results on the Beer review dataset after adding continuous constraint. Top-$10\%$ tokens are selected for the test datasets. Causal rationalization performs the best in all aspects in terms of capturing human annotations.}
\vskip -0.05in
\resizebox{\columnwidth}{!}{%
\begin{tabular}{lcccccccccccc} 
\toprule
\multirow{2}{*}{ Methods } & \multicolumn{4}{c}{ Appearance } & \multicolumn{4}{c}{ Aroma } & \multicolumn{4}{c}{ Palate } \\
\cmidrule(lr){2-5}
\cmidrule(lr){6-9}
\cmidrule(lr){10-13}
& Acc $(\uparrow)$ & P $(\uparrow)$ &  R $(\uparrow)$ & F1 $(\uparrow)$ & Acc $(\uparrow)$  & P $(\uparrow)$ &R $(\uparrow)$ & F1 $(\uparrow)$ & Acc $(\uparrow)$  & P $(\uparrow)$& R $(\uparrow)$& F1 $(\uparrow)$ \\
\midrule
VIB & $\mathbf{93.8(2.4)}$  &$52.6(2.0)$ &$26.0(2.3)$ & $32.9(2.1)$ & $85.0(1.1)$ &$54.2(2.9)$ &$31.6(1.9)$ & $37.7(2.8)$ & $81.5(3.2)$  &$ 41.2(2.1)$ & $ 35.1(3.0)$ & $35.2(2.8) $  \\
RNP & $91.5(1.7)$ &$40.0(1.4)$ &$20.3(1.9)$ & $25.2(1.7)$ & $84.0(2.1) $ &$ 49.1(3.2)$ &$28.7(2.2) $ & $ 32.0(2.5)$ & $80.3(3.4) $ &$38.6(1.8) $ &$31.1(2.3) $ & $29.7(2.0) $  \\
FR & $93.5(1.0)$  & $51.9(1.1)$&$25.1(2.0)$ & $31.8(1.6)$ & $88.0(1.8)$ &$54.8(3.5)$ &$33.7(2.6)$ & $39.5(3.7)$ & $ 82.0(2.1)$ & $ 44.3(3.4)$ & $32.5(2.7) $& $ 33.7(3.1)$  \\
A2R & $91.5(2.2)$  & $55.0(0.8)$&$25.8(1.6)$ & $34.3(1.4)$ & $85.5(1.9)$ &$61.3(2.8)$ &$34.8(3.1)$ & $41.2(3.3)$ & $80.5(2.4)$ & $40.1(2.9)$ & $34.2(3.2)$& $34.6(3.2)$  \\
INVRAT &  $91.0(3.1)$ & $56.4(2.5)$ & $27.3(1.2)$ & $36.7(2.1)$ &$\mathbf{90.0(3.0)}$ & $49.6(3.1)$ & $27.5(1.9)$ & $33.2(2.6)$ & $80.0(1.8)$ & $42.2(3.2)$ & $32.2(1.6)$ & $31.9(2.4)$\\
CR(ours)  & $92.4(1.7)$ &$\mathbf{59.7(1.9)}$ &$\mathbf{31.6(1.6)}$ & $\mathbf{39.0(1.5)}$  &  $86.5(2.1)$ &$\mathbf{68.0(2.9)}$  &$\mathbf{42.0(3.0)}$& $\mathbf{49.1(2.8)}$&  $\mathbf{82.5(2.3)} $ & $\mathbf{44.7(2.5)} $ & $\mathbf{37.3(2.0)} $& $\mathbf{38.1(2.1)} $ \\
\bottomrule
\end{tabular}
}
\label{beer conti}
\end{table*}

\begin{table*}[t] 
\centering
\caption{Results on the Hotel review dataset. Top-$10\%$ tokens are highlighted for evaluation. Causal rationalization performs the best in all aspects in terms of capturing human annotations.}
\vskip -0.1in
\small
\begin{tabular}{lcccc} 
\toprule
\multirow{2}{*}{ Methods }  & \multicolumn{4}{c}{ Location }\\
 &Acc $(\uparrow)$  & P $(\uparrow)$& R $(\uparrow)$ & F1 $(\uparrow)$\\
\midrule
VIB  & $93.3(1.8)$ &$38.3(4.1)$ &$41.6(6.4)$ & $35.3(4.7)$  \\
RNP  & $94.9(1.7)$ &$37.2(2.1)$ &$39.8(3.3)$ & $34.0(3.9)$  \\
FR & $\mathbf{97.3(1.8)}$ &$35.5(1.7)$ &$40.6(1.3)$ & $33.5(1.2)$  \\
A2R & $92.0(2.2)$ &$37.8(2.9)$ &$40.1(2.1)$ & $34.4(3.2)$  \\
CR(ours)  & $94.0(2.1)$ &$\mathbf{39.4(1.0)}$ &$\mathbf{44.2(1.5)}$ & $\mathbf{36.9(1.0)}$  \\
\bottomrule
\end{tabular}
\vskip -0.1in
\label{hotel-result-table}
\end{table*}

\subsection{Hotel Review Results}
\label{hotel-result}
Since Hotel review data have fewer continuous rationales, we compare all the baseline methods with CR without the continuity constraint. We don't include INVART because for Hotel review data, it doesn't has continuous scores and we can't follow the same way of selecting environments as Beer review data. From Table \ref{hotel-result-table}, upon conducting our analysis, we have observed that our approach outperforms other baseline methods in terms of capturing human annotations. Based on these results, our method can be considered for long-text data.

\subsection{Sensitivity Analyses}
\begin{figure}[!htbp]
\centering
        \begin{subfigure}
        \centering
            \includegraphics[width=0.4\columnwidth]{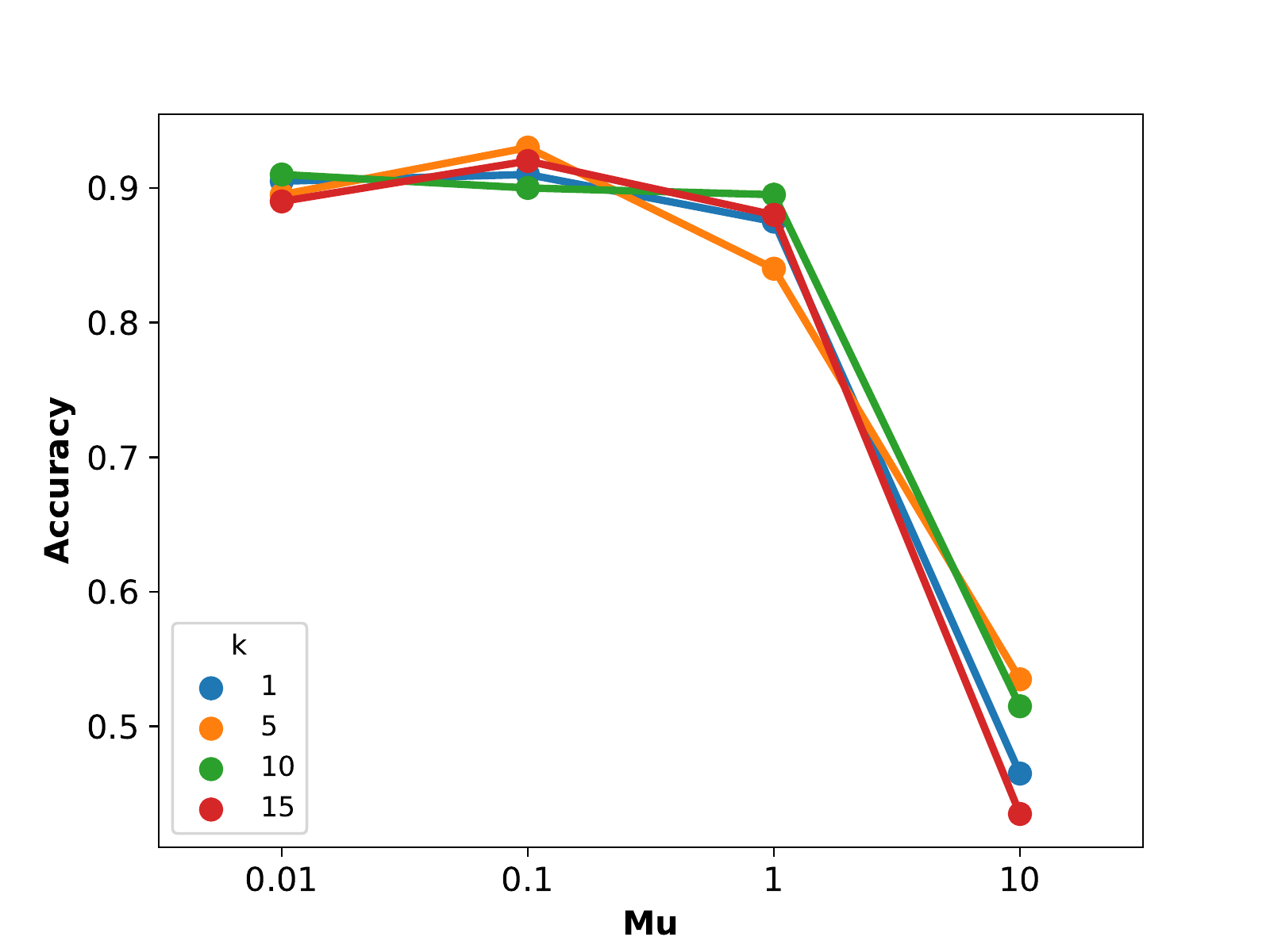}
            \includegraphics[width=0.4\columnwidth]{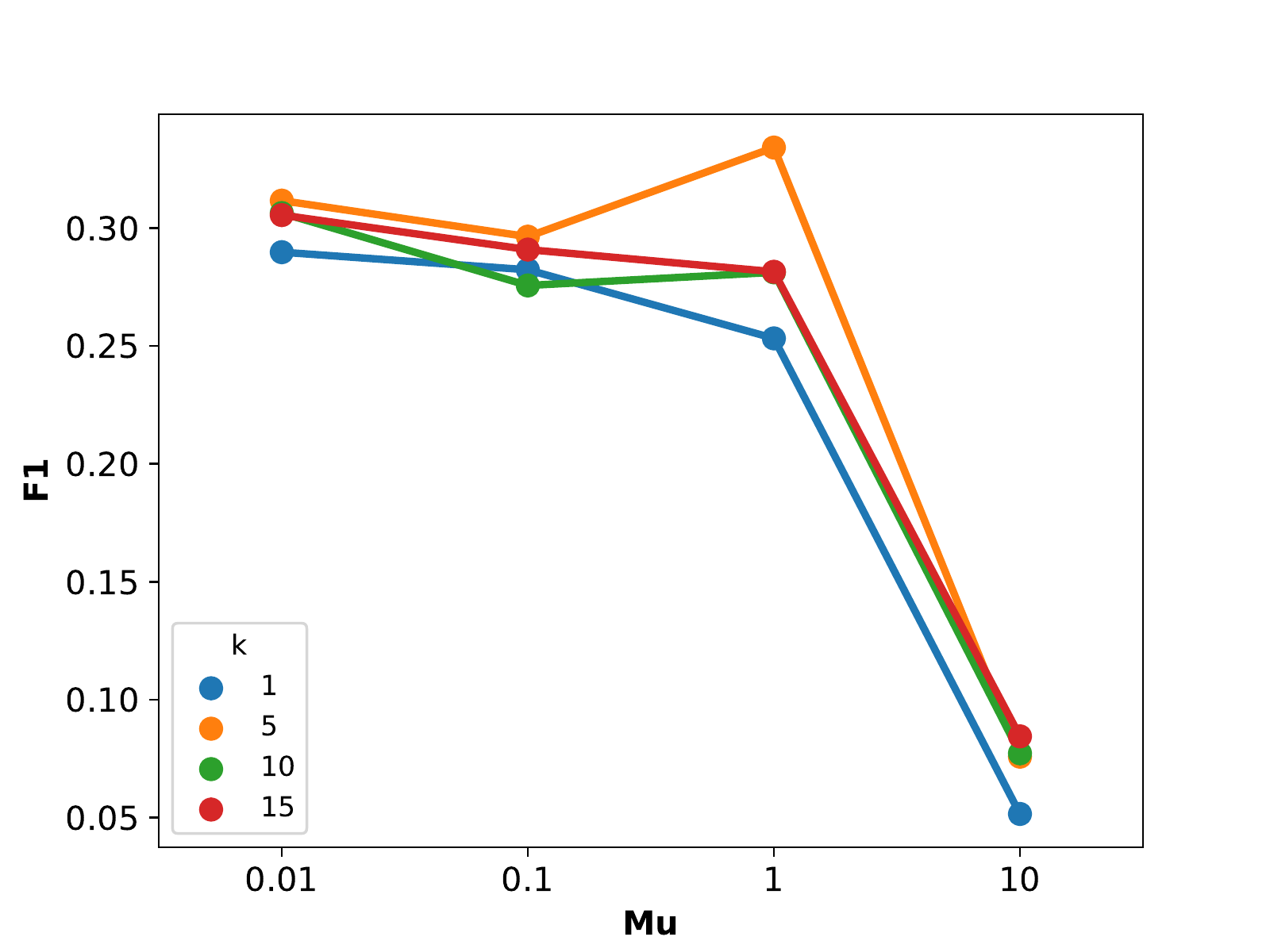}
            \caption{Accuracy and F1 score for Appearance.}
                    \label{sensi-appearance}
        \end{subfigure}
    
    \begin{subfigure}
        \centering
            \includegraphics[width=0.4\columnwidth]{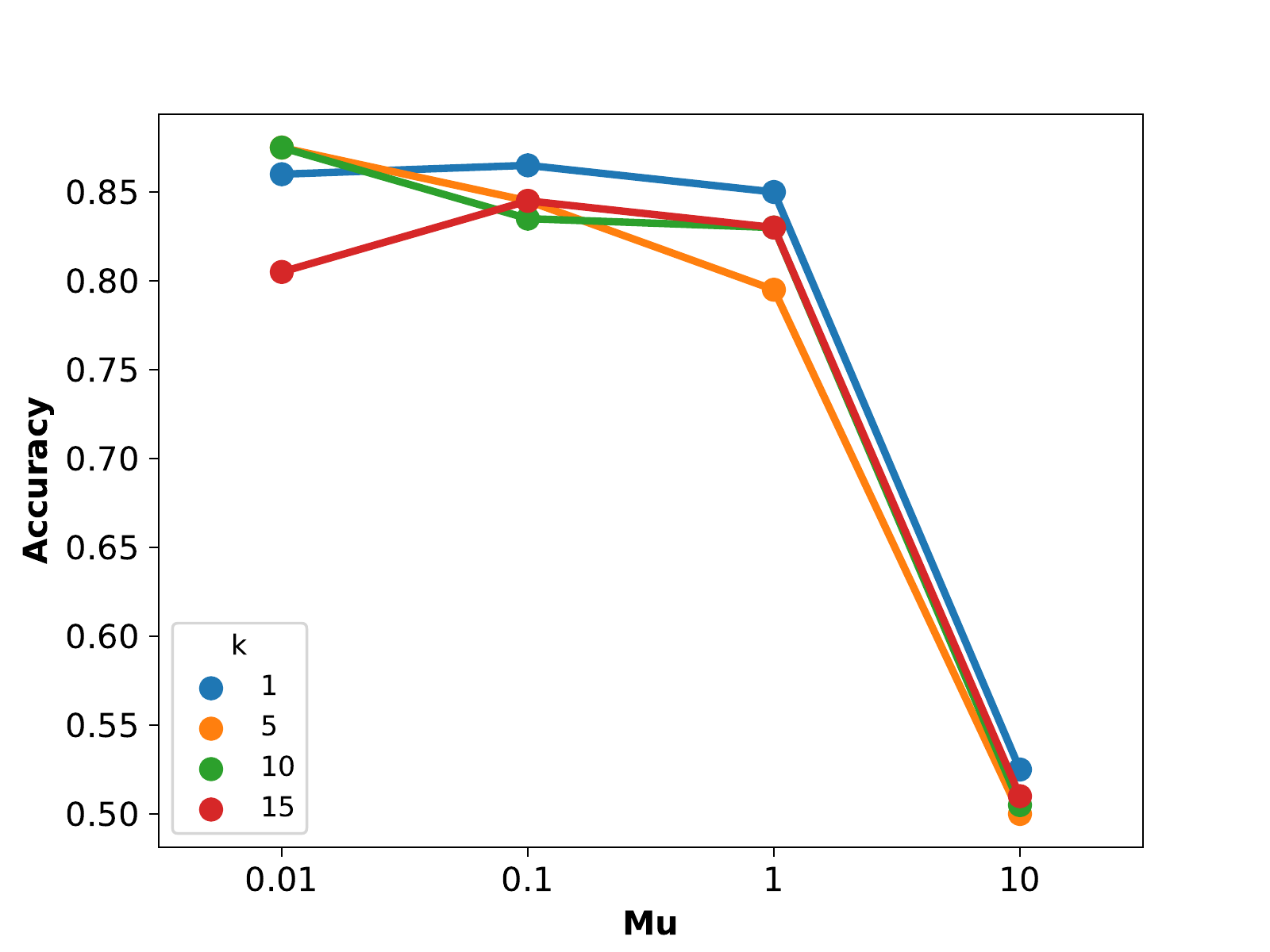}
            \includegraphics[width=0.4\columnwidth]{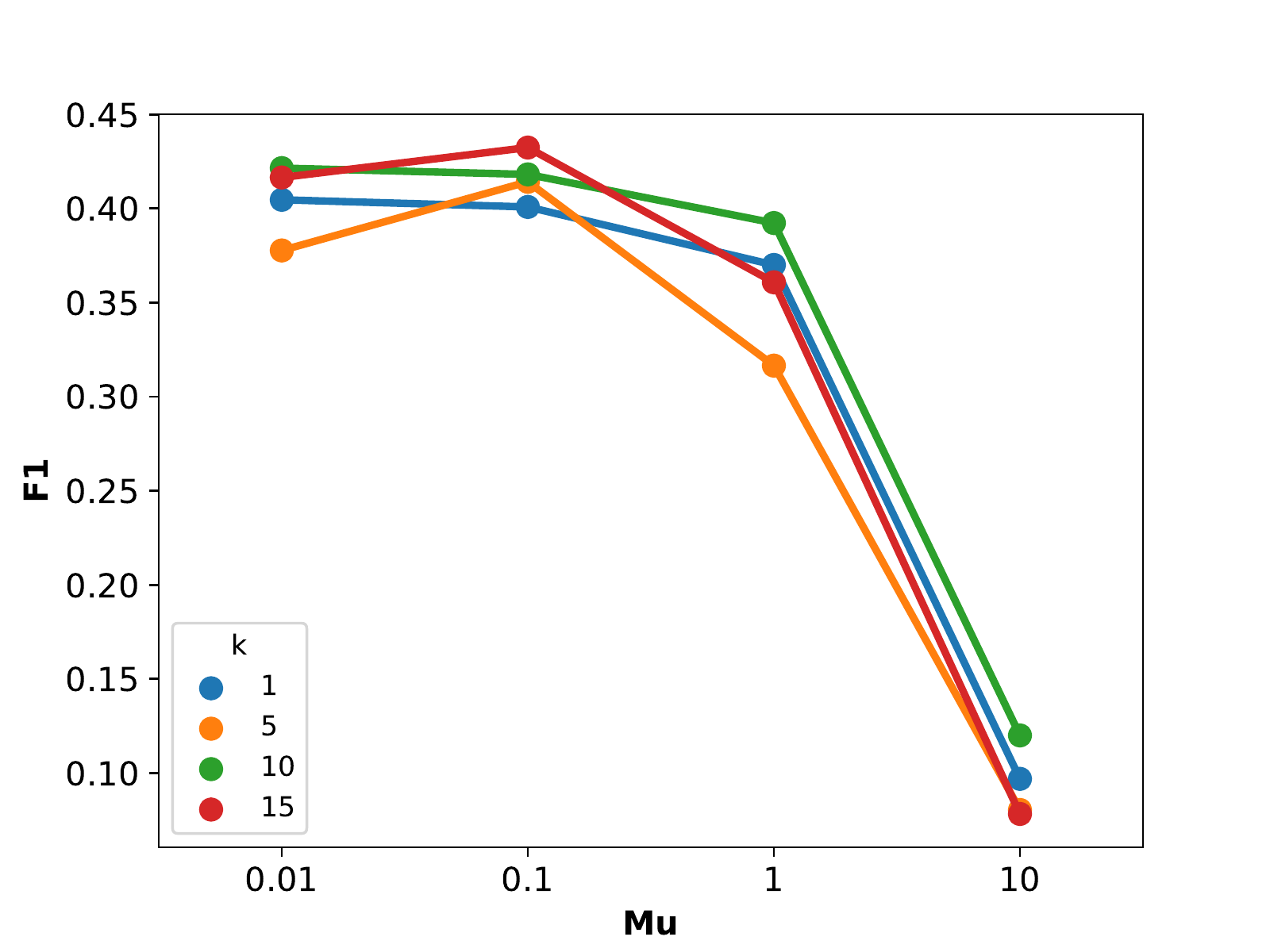}  
        \caption{Accuracy and F1 score for Aroma.}
                 \label{sensi-aroma}
        \end{subfigure}

            \begin{subfigure}
        \centering
            \includegraphics[width=0.4\columnwidth]{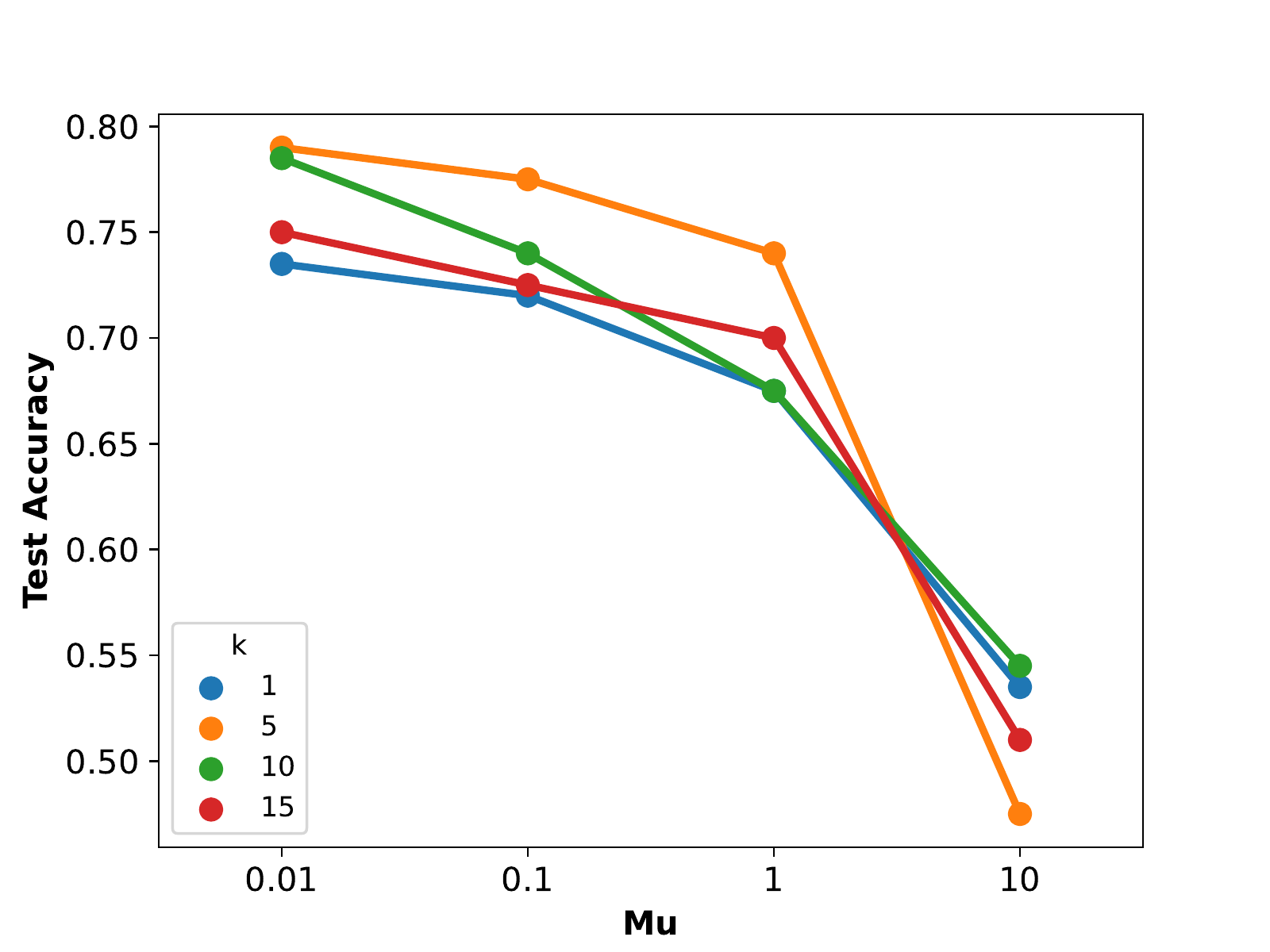}
            \includegraphics[width=0.4\columnwidth]{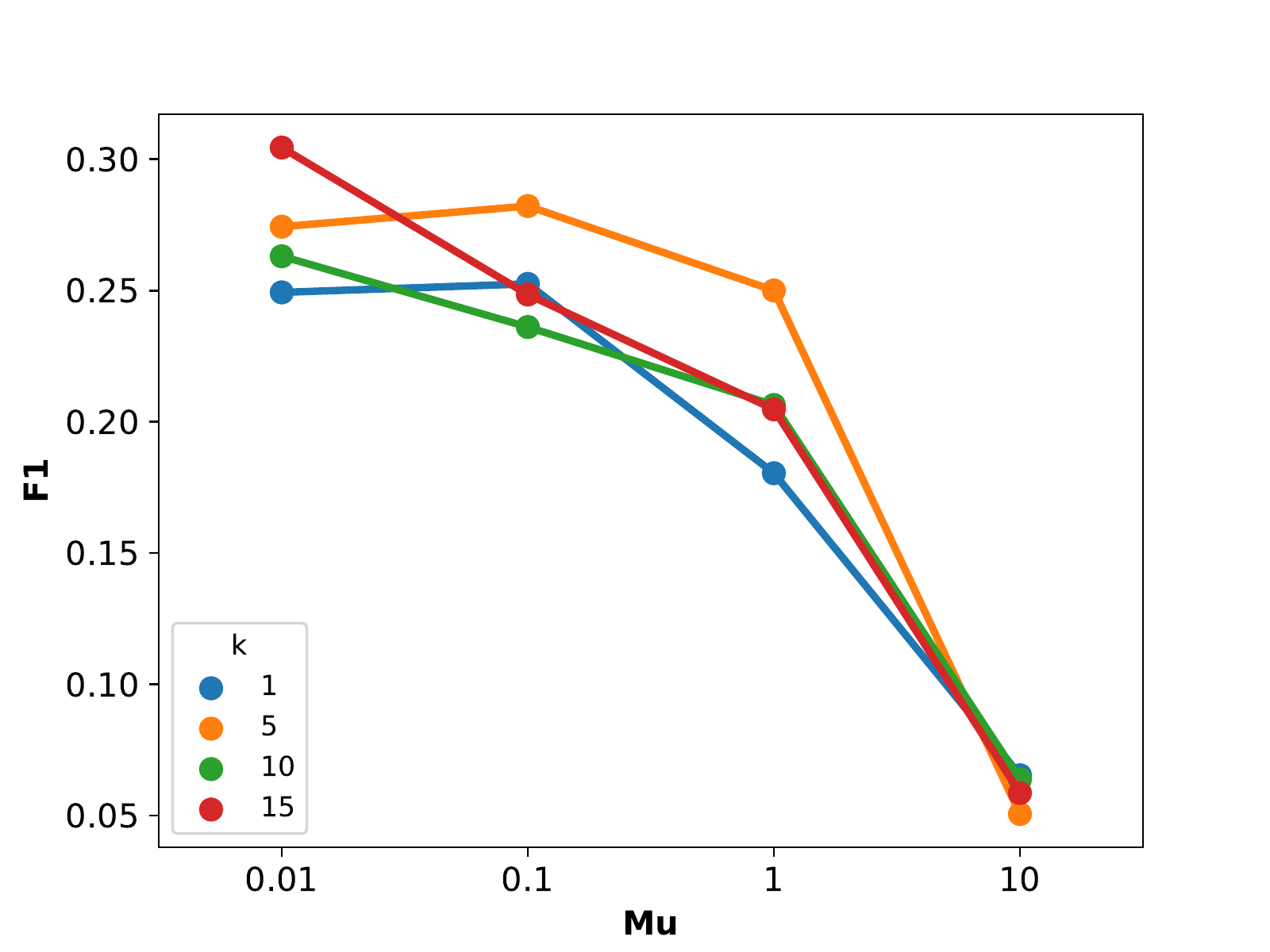}  
     \caption{Accuracy and F1 score for Palate.}
                      \label{sensi-palate}
        \end{subfigure}
        \label{sensitivity}
\end{figure}

In the previous experiments, we set $\mu=0.1$ and $k=5\%$. To understand the sensitivity of the two parameters, we re-run the experiments on real Beer review data, with $\mu=\{0.01,0.1,1,10\}$ and $k=\{1\%,5\%,10\%,15\%\}$ while keeping the sparsity constraint to be $0.1$. We select Top-$10\%$ tokens and use accuracy and F1 for the evaluation. Figure \ref{sensi-appearance},\ref{sensi-aroma}, and \ref{sensi-palate} summarize our results. It can be seen that our causal rationalization approach's performance is not sensitive to $k$ and $\mu$ when $\mu$ is not too large e.g. $(0.01,0.1,1)$.

\subsection{Visualization Examples}
\subsubsection{Beer review}\label{all Visualization beer Results}
We provide three examples for each aspect in terms of all the methods in \cref{vis beer examples}.

\subsubsection{GA}
\label{all Visualization ga Results}

Since we are more concerned about positive patients who are diagnosed with GA, we present one example of them here. We have converted the medical codes to their corresponding descriptions. As the descriptions can be quite lengthy, we have only included selected codes. In instances where there are multiple consecutive codes, we have only displayed one.

Compared to the rationales found by baseline methods, the ones predicted by our proposed method hit more risk factors of GA.
As shown in the first column of Table 9, the patient suffered eyesight defect (BILATERAL FIELD DETECT), irregular heart beat (ATRIAL FIBRILLATION), and diabetes (TYPE I DIABETES WITH DIABETIC POLYNEUROPATHY), all of which are clinically associated with GA as strong risk factors.
In comparison, many of the rationales returned by baseline methods are clinically irrelevant to GA, hence less robust.

It is essential to consider the importance of model interpretability in high-stakes applications like healthcare. Although predictive accuracy is a vital aspect, the capacity of a model to provide self-explanatory and interpretable predictions is paramount in fostering trust among healthcare professionals. Furthermore, regulatory compliance necessitates interpretable models, as authorities may mandate explainability to ensure safety, efficacy, and fairness in healthcare algorithms. Given the domain-specific demands for algorithms in healthcare, model interpretability often takes precedence over predictive accuracy, as long as the accuracy is on par with other less interpretable algorithms. Consequently, we argue that the negligible discrepancy in clinical outcome predictions between our proposed method and baselines should be considered within the context of the critical role of interpretability in healthcare applications.

\begin{table}[!htp]
\centering
    \caption{Visualization examples of Beer review data.}
    \resizebox{1\textwidth}{!}{
    \begin{minipage}{1.2\textwidth}
            
    \begin{tabular}{ p{4.8cm}  p{4.8cm}  p{4.8cm}  p{4.8cm}}
    
        \toprule
\textbf{CR}      
& \textbf{VIB}  
& \textbf{RNP}
& \textbf{FR}
\\\midrule
\textbf{Aspect}: Appearance
\newline
\textbf{Label}: Positive
\newline
\textbf{Pred}: Positive
\newline
poured from bottle into shaker pint glass . \redulns{a :} \redulns{pale tan-yellow} \blackulns{with} \redulns{very thin white head} \blackulns{that} \redulns{quickly disappears} . s : malt , caramel ... and skunk t : very bad . hint of malt . mostly just tastes bad though . m : acidic , thin , and watery d : not drinkable . unbelievably bad . stay away ... you have several thousand other beers that taste better to spend you money on . do n't be a fool like me . 
& 
\textbf{Aspect}: Appearance 
\newline
\textbf{Label}: Positive
\newline
\textbf{Pred}: Positive
\newline
poured from bottle into shaker pint \textcolor{blue}{glass} . \blackulns{a :}
\blueulns{pale tan-yellow} \blackulns{with very} \blueulns{thin white} \blackulns{head that} \blueulns{quickly disappears} . s : malt , caramel ... and skunk t : very \textcolor{blue}{bad} . hint of malt . mostly just tastes bad though . m : acidic , \textcolor{blue}{thin} , and \textcolor{blue}{watery} d : not drinkable . unbelievably bad . stay away ... you have several thousand other beers that taste better to spend you money on . do n't be a fool like me .      
& 
\textbf{Aspect}: Appearance 
\newline
\textbf{Label}: Positive
\newline
\textbf{Pred}: Positive
\newline
poured from bottle into shaker pint glass . \blackulns{a :} \orangeulns{pale tan-yellow} \blackulns{with very} \orangeulns{thin} \blackulns{white head that quickly} \orangeulns{disappears} . s : malt , caramel ... and skunk t : very bad . hint of malt . mostly just tastes bad though . m : acidic , thin , and \textcolor{orange}{watery} d : \textcolor{orange}{not} drinkable . unbelievably bad . stay away ... you have several thousand other \textcolor{orange}{beers} that taste better to spend you money on . do n't be a fool like me .     
&
\textbf{Aspect}: Appearance 
\newline
\textbf{Label}: Positive
\textbf{Pred}: Positive
\newline
poured from bottle into shaker pint \textcolor{green}{glass} . \blackulns{a :} \greenulns{pale tan-yellow} \blackulns{with very} \greenulns{thin} \blackulns{white head that quickly} \greenulns{disappears} \textcolor{green}{.} s : malt , caramel ... and skunk t : very bad . hint of malt . mostly just tastes \textcolor{green}{bad} though . m : acidic , thin , and \textcolor{green}{watery} d : \textcolor{green}{not} drinkable . unbelievably \textcolor{green}{bad} . stay away ... you have several thousand other beers that taste better to spend you money on . do n't be a fool like me .   
\\\hline
\textbf{Aspect}: Aroma
\newline
\textbf{Label}: Negative
\newline
\textbf{Pred}: Negative
\newline
medium head that quickly disappears . lacing is spotty . \blackulns{the} \redulns{smell is rancid . it smells like swamp} \textcolor{red}{gas} . i am going to assume it is from the can but with miller , who knows ? the best way to describe this brew is `` sugar water with a slight watermelon taste '' . this is a very \textcolor{red}{sweet} tasting beer hence it 's gloss on the can `` the champagne of beers ! '' . overall not bad for a macro but \textcolor{red}{nothing} exciting going on here . notes : shared with my old man at the kitchen table . he buys whatever is \textcolor{red}{cheapest} and i take the opportunity to review \textcolor{red}{bad} beer . i look at it this way . i wish the old man would buy better beer , but i get to review beer i normally would n't buy anyways .
&
\textbf{Aspect}: Aroma
\newline
\textbf{Label}: Negative
\newline
\textbf{Pred}: Negative
\newline
medium head that quickly disappears . lacing is spotty . \blackulns{the} \blackulns{smell} \blackulns{is rancid . it} \blueulns{smells like} \blackulns{swamp} gas . i am going to assume \textcolor{blue}{it is from} the can but \textcolor{blue}{with} miller , who \textcolor{blue}{knows} ? the best way to describe this brew is `` sugar water with a slight watermelon taste '' . this is a very sweet tasting beer hence it 's gloss on the can `` the champagne of beers ! '' . overall not bad for a macro but nothing exciting going on here . notes : shared with my old man at the kitchen table . he buys whatever is cheapest and i take the opportunity to review bad beer . i look at it this way . i wish the \textcolor{blue}{old man would buy better} beer , but i get to review beer i normally would n't buy anyways .
&
\textbf{Aspect}: Aroma
\newline
\textbf{Label}: Negative
\newline
\textbf{Pred}: Negative
\newline
medium head that quickly disappears . lacing is spotty . \blackulns{the} \orangeulns{smell} \blackulns{is rancid . it} \orangeulns{smells like swamp} \textcolor{orange}{gas} . i am going to assume it is from the can but with miller , who knows ? the best way to describe this brew is \textcolor{orange}{``} sugar water with a slight \textcolor{orange}{watermelon taste}'' . this is a very sweet tasting beer hence it 's gloss on the can `` the champagne of beers ! '' . overall not \textcolor{orange}{bad} for a macro but \textcolor{orange}{nothing} exciting going on here . notes : shared with my old man at the kitchen table . he buys whatever is cheapest and i take the opportunity to review bad beer . i look at it this way . i wish the old man would buy better beer , but i get to review beer i normally would n't buy anyways .
&
\textbf{Aspect}: Aroma
\newline
\textbf{Label}: Negative
\newline
\textbf{Pred}: Negative
\newline
medium head that quickly disappears . lacing is spotty . \blackulns{the} \greenulns{smell} \blackulns{is} \greenulns{rancid} \blackulns{. it smells like swamp} \textcolor{green}{gas} . i am going to assume it is from the can but with \textcolor{green}{miller} , who \textcolor{green}{knows} ? the best way to describe this brew is `` sugar water with a \textcolor{green}{slight} watermelon taste '' . this is a very sweet tasting beer hence it 's gloss on the can `` the champagne of beers ! '' . overall \textcolor{green}{not bad} for a macro but nothing exciting going on here . notes : shared with my old man at the kitchen table . he buys whatever is \textcolor{green}{cheapest} and i take the opportunity to review \textcolor{green}{bad} beer . i look at it this way . i wish the old man would buy \textcolor{green}{better} beer , but i get to review beer i normally would n't buy anyways .
\\\hline
\textbf{Aspect}: Palate
\newline
\textbf{Label}: Positive
\newline
\textbf{Pred}: Positive
\newline
 sparkling yellow hue with large marshmallow head . light hops , corn , alcohol , grass in the nose . grass notes , bready , soapy hops in the finish . \redulns{smooth} \blackulns{but} \redulns{slick} \blackulns{in} \redulns{mouthfeel} . \textcolor{red}{highly} drinkable and \textcolor{red}{enjoyable} . sierra nevada is still the kings of hops even though this beer is a more average brew for them .
 &
 \textbf{Aspect}: Palate
\newline
\textbf{Label}: Positive
\newline
\textbf{Pred}: Positive
\newline
 sparkling yellow hue with large marshmallow head . light hops , corn , alcohol , grass in the nose . grass notes , bready , soapy hops in the finish . \blueulns{smooth} but \blueulns{slick} in mouthfeel . \textcolor{blue}{highly} drinkable and \textcolor{blue}{enjoyable} . \textcolor{blue}{sierra} nevada is still the kings of hops even though this beer is a more \textcolor{blue}{average} brew for them .
 &
 \textbf{Aspect}: Palate
\newline
\textbf{Label}: Positive
\newline
\textbf{Pred}: Positive
\newline
 sparkling yellow hue with large marshmallow head . light hops , corn , alcohol , grass in the nose . \textcolor{orange}{grass} notes , bready , soapy hops in the finish . \blackulns{smooth} \orangeulns{but} \blackulns{slick in mouthfeel} . highly drinkable and enjoyable . sierra nevada is still the \textcolor{orange}{kings of hops} even though this beer is a more average brew for them .
 &
 \textbf{Aspect}: Palate
\newline
\textbf{Label}: Positive
\newline
\textbf{Pred}: Positive
\newline
 sparkling yellow hue with large \textcolor{green}{marshmallow} head . light \textcolor{green}{hops} , corn , alcohol , grass in the nose . grass notes , bready , soapy hops in the finish . \greenulns{smooth} \blackulns{but slick in mouthfeel} . \textcolor{green}{highly} drinkable and \textcolor{green}{enjoyable} . sierra nevada is still the kings of hops even though this beer is a more average brew for them .\\

\bottomrule
\label{vis beer examples}
    \end{tabular}
        \end{minipage}
    }
    
\end{table}

\begin{table}[!ht]
\centering
    \caption{Visualization examples of two GA patients.
    }
    \begin{minipage}{1\textwidth}
        \footnotesize
            \centering
    \begin{tabular}{ p{3.9cm}  p{3.9cm}  p{3.9cm}  p{3.9cm}}
        \toprule
\textbf{CR}      
& \textbf{VIB}  
& \textbf{RNP}
& \textbf{FR}
\\\midrule
\textbf{Label}: Positive
\newline

\textbf{Pred}: Positive
\newline

\textcolor{blue}{\XSolidBold} \quad  HEMORRHAGE, NOT ELSEWHERE CLASSIFIED 
\newline

\textcolor{blue}{\XSolidBold} \quad PAIN IN RIGHT HIP  
\newline

\textcolor{red}{\CheckmarkBold} \quad HOMONYMOUS BILATERAL FIELD DEFECTS, LEFT SIDE 
\newline 

\textcolor{red}{\CheckmarkBold} \quad  CHRONIC ATRIAL FIBRILLATION 
\newline

\textcolor{red}{\CheckmarkBold} \quad HOMONYMOUS BILATERAL FIELD DEFECTS, UNSPECIFIED SIDE
\newline

\textcolor{red}{\CheckmarkBold} \quad  OTHER OPTIC ATROPHY, RIGHT EYE
\newline

\textcolor{red}{\CheckmarkBold} \quad  TYPE 1 DIABETES MELLITUS WITH DIABETIC POLYNEUROPATHYSIDE
\newline

\textcolor{blue}{\XSolidBold} \quad  ERECTILE DYSFUNCTION DUE TO ARTERIAL INSUFFICIENCYSIDE
\newline

\textcolor{red}{\CheckmarkBold} \quad TYPE 1 DIABETES MELLITUS WITH DIABETIC POLYNEUROPATHY
\newline

\textcolor{blue}{\XSolidBold} \quad  ACQUIRED KERATOSIS [KERATODERMA] PALMARIS ET PLANTARIS

&
\textbf{Label}: Positive
\newline

\textbf{Pred}: Positive
\newline  

\textcolor{blue}{\XSolidBold} \quad ADVERSE EFFECT OF ANTICOAGULANTS, INITIAL ENCOUNTER 
\newline

\textcolor{blue}{\XSolidBold} \quad HEMORRHAGE, NOT ELSEWHERE CLASSIFIED
\newline

\textcolor{blue}{\XSolidBold} \quad DEHYDRATION
\newline

\textcolor{blue}{\XSolidBold} \quad SUBLUXATION COMPLEX (VERTEBRAL) OF LUMBAR REGION  
\newline

\textcolor{red}{\CheckmarkBold} \quad  HOMONYMOUS BILATERAL FIELD DEFECTS, LEFT SIDE
\newline

\textcolor{red}{\CheckmarkBold} \quad  CHRONIC ATRIAL FIBRILLATION
\newline

\textcolor{red}{\CheckmarkBold} \quad HOMONYMOUS BILATERAL FIELD DEFECTS, UNSPECIFIED SIDE         \newline

\textcolor{blue}{\XSolidBold} \quad SUBLUXATION COMPLEX (VERTEBRAL) OF LUMBAR REGION            \newline

\textcolor{red}{\CheckmarkBold} \quad TYPE 1 DIABETES MELLITUS WITHOUT COMPLICATIONS
\newline

\textcolor{blue}{\XSolidBold} \quad  STRAIN OF MUSCLE(S) AND TENDON(S) OF THE ROTATOR CUFF OF RIGHT SHOULDER, INITIAL ENCOUNTER
\newline

\textcolor{blue}{\XSolidBold} \quad  XEROSIS CUTIS             
\newline

\textcolor{blue}{\XSolidBold} \quad  PERIPHERAL VASCULAR DISEASE, UNSPECIFIED||  
\newline

& 
\textbf{Label}: Positive
\newline

\textbf{Pred}: Positive
\newline

\textcolor{blue}{\XSolidBold} \quad ALCOHOL ABUSE WITH INTOXICATION, UNSPECIFIED
\newline

\textcolor{blue}{\XSolidBold} \quad  PAIN IN RIGHT HIP
\newline

\textcolor{blue}{\XSolidBold} \quad  SUBLUXATION COMPLEX (VERTEBRAL) OF LUMBAR REGION
\newline

\textcolor{blue}{\XSolidBold} \quad  PAIN IN RIGHT SHOULDER
\newline

\textcolor{blue}{\XSolidBold} \quad  LOW BACK PAIN
\newline

\textcolor{red}{\CheckmarkBold} \quad UNSPECIFIED ATRIAL FIBRILLATION
\newline

\textcolor{blue}{\XSolidBold} \quad STRAIN OF MUSCLE(S) AND TENDON(S) OF THE ROTATOR CUFF OF RIGHT SHOULDER, INITIAL ENCOUNTER
\newline

\textcolor{blue}{\XSolidBold} \quad LOW BACK PAIN
\newline

\textcolor{red}{\CheckmarkBold} \quad  TYPE 1 DIABETES MELLITUS WITH DIABETIC POLYNEUROPATHY
\newline

\textcolor{blue}{\XSolidBold} \quad STRAIN OF MUSCLE(S) AND TENDON(S) OF THE ROTATOR CUFF OF RIGHT SHOULDER, INITIAL ENCOUNTER
\newline
                          
\textcolor{blue}{\XSolidBold} \quad PERIPHERAL VASCULAR DISEASE, UNSPECIFIED
\newline

\textcolor{red}{\CheckmarkBold} \quad  UNSPECIFIED ATRIAL FIBRILLATION
\newline
    
&
\textbf{Label}: Positive
\newline

\textbf{Pred}: Positive
\newline

\textcolor{blue}{\XSolidBold} \quad HEMATURIA, UNSPECIFIED
\newline

\textcolor{blue}{\XSolidBold} \quad ADVERSE EFFECT OF ANTICOAGULANTS, INITIAL ENCOUNTER
\newline

\textcolor{red}{\CheckmarkBold} \quad CHRONIC FATIGUE, UNSPECIFIED
\newline

\textcolor{blue}{\XSolidBold} \quad SUBLUXATION COMPLEX (VERTEBRAL) OF LUMBAR REGION
\newline

\textcolor{blue}{\XSolidBold} \quad STRAIN OF MUSCLE, FASCIA AND TENDON OF LOWER BACK, INITIAL ENCOUNTER
\newline

\textcolor{red}{\CheckmarkBold} \quad HOMONYMOUS BILATERAL FIELD DEFECTS, LEFT SIDE
\newline

\textcolor{blue}{\XSolidBold} \quad SUBLUXATION COMPLEX (VERTEBRAL) OF LUMBAR REGION
\newline

\textcolor{blue}{\XSolidBold} \quad BARRETT'S ESOPHAGUS WITH DYSPLASIA, UNSPECIFIED
\newline

\textcolor{blue}{\XSolidBold} \quad STRAIN OF MUSCLE(S) AND TENDON(S) OF THE ROTATOR CUFF OF RIGHT SHOULDER, INITIAL ENCOUNTER
\newline

\textcolor{red}{\CheckmarkBold} \quad TYPE 1 DIABETES MELLITUS WITH DIABETIC POLYNEUROPATHY
\newline

\textcolor{blue}{\XSolidBold} \quad PAIN IN LEFT SHOULDER
\newline

\textcolor{blue}{\XSolidBold} \quad ACQUIRED KERATOSIS [KERATODERMA] PALMARIS ET PLANTARIS
\newline

\textcolor{blue}{\XSolidBold} \quad UNILATERAL PRIMARY OSTEOARTHRITIS OF FIRST CARPOMETACARPAL JOINT, LEFT HAND  \\

\bottomrule
\label{vis ga examples}
    \end{tabular}
        \end{minipage}
\end{table}

\clearpage

\section{Proof of Theorem \ref{theorem}}
\label{proof}
\subsection{Identification of CPNS}
    
\noindent \textbf{Proof:} Let's denote the logical operators \textit{and}, \textit{or} as $\wedge, \vee$, respectively. Given $\boldsymbol{X}= \boldsymbol{x}$, firstly we know 
\begin{equation}
\left\{Y\left(Z_i = z_i,\boldsymbol{Z}_{-i}= \boldsymbol{z}_{-i}\right)=y\right\} \vee\left\{Y\left(Z_i = z_i,\boldsymbol{Z}_{-i}= \boldsymbol{z}_{-i},\right) \neq y\right\}= \text{True}.
\label{1}
\end{equation}

Then we have
\begin{eqnarray}
&&\left\{Y\left(Z_j \neq z_j,\boldsymbol{Z}_{-j}= \boldsymbol{z}_{-j}\right)=y\right\}\nonumber \\
\stackrel{\eqref{1}}{=}&&\left\{Y\left(Z_j \neq z_j,\boldsymbol{Z}_{-j}= \boldsymbol{z}_{-j}\right)=y\right\}  \nonumber\\
&& \wedge 
\left[\left\{Y\left(Z_j=z_j, \boldsymbol{Z}_{-j}=\boldsymbol{z}_{-j},\boldsymbol{X}= \boldsymbol{x}\right)=y\right\} \vee\left\{Y\left(Z_j=z_j, \boldsymbol{Z}_{-j}=\boldsymbol{z}_{-j}\right) \neq y\right\}\right]\nonumber \\
=&&\left[\left\{Y\left(Z_j \neq z_j,\boldsymbol{Z}_{-j}= \boldsymbol{z}_{-j}\right)=y\right\}\wedge \left\{Y\left(Z_j=z_j, \boldsymbol{Z}_{-j}=\boldsymbol{z}_{-j}\right)=y\right\}\right] \nonumber\\
&&\vee\left[\left\{Y\left(Z_j \neq z_j,\boldsymbol{Z}_{-j}= \boldsymbol{z}_{-i}\right)=y\right\}\wedge \left\{Y\left(Z_j=z_j, \boldsymbol{Z}_{-j}=\boldsymbol{z}_{-j}\right) \neq y\right\}\right] \nonumber\\
\stackrel{\eqref{monotonicity}}{=} && \left[\left\{Y\left(Z_j \neq z_j,\boldsymbol{Z}_{-j}= \boldsymbol{z}_{-j}\right)=y\right\}\wedge \left\{Y\left(Z_j=z_j, \boldsymbol{Z}_{-j}=\boldsymbol{z}_{-j}\right)=y\right\}\right], ~~~~
\label{2}
\end{eqnarray}
where we use the monotonicity assumption in \eqref{monotonicity}.

Also, we know 
\begin{equation}
\left\{Y\left(Z_i \neq z_i,\boldsymbol{Z}_{-i}= \boldsymbol{z}_{-i},\right)=y\right\} \vee\left\{Y\left(Z_i \neq z_i,\boldsymbol{Z}_{-i}= \boldsymbol{z}_{-i}\right) \neq y\right\}= \text{True}.
\label{3}
\end{equation}

Then we can get
\begin{eqnarray}
&&\left\{Y\left(Z_j = z_j,\boldsymbol{Z}_{-j}= \boldsymbol{z}_{-j}\right)=y\right\} \nonumber \\
\stackrel{\eqref{3}}{=}&&\left\{Y\left(Z_j = z_j,\boldsymbol{Z}_{-j}= \boldsymbol{z}_{-j}\right)=y\right\}  \nonumber\\
&& \wedge 
\left[\left\{Y\left(Z_j\neq z_j, \boldsymbol{Z}_{-j}=\boldsymbol{z}_{-j}\right)=y\right\} \vee\left\{Y\left(Z_j\neq z_j, \boldsymbol{Z}_{-j}=\boldsymbol{z}_{-j}\right) \neq y\right\}\right]\nonumber \\
=&&\left[\left\{Y\left(Z_j = z_j,\boldsymbol{Z}_{-j}= \boldsymbol{z}_{-j}\right)=y\right\}\wedge \left\{Y\left(Z_j\neq z_j, \boldsymbol{Z}_{-j}=\boldsymbol{z}_{-j}\right)=y\right\}\right] \nonumber\\
&&\vee\left[\left\{Y\left(Z_j = z_j,\boldsymbol{Z}_{-j}= \boldsymbol{z}_{-i}\right)=y\right\}\wedge \left\{Y\left(Z_j\neq z_j, \boldsymbol{Z}_{-j}=\boldsymbol{z}_{-j}\right) \neq y\right\}\right] \nonumber\\
\stackrel{\eqref{2}}{=} && \left\{Y\left(Z_j \neq z_j,\boldsymbol{Z}_{-j}= \boldsymbol{z}_{-j}\right)=y\right\} \nonumber\\
&&\vee\left[\left\{Y\left(Z_j = z_j,\boldsymbol{Z}_{-j}= \boldsymbol{z}_{-i}\right)=y\right\}\wedge \left\{Y\left(Z_j\neq z_j, \boldsymbol{Z}_{-j}=\boldsymbol{z}_{-j}\right) \neq y\right\}\right]. \nonumber\\
\label{4}
\end{eqnarray}
Based on the consistency assumption in  \eqref{consistency}, we either have $\left\{Y\left(Z_j \neq z_j,\boldsymbol{Z}_{-j}= \boldsymbol{z}_{-j}\right)=y\right\}$ or $\left\{Y\left(Z_j\neq z_j, \boldsymbol{Z}_{-j}=\boldsymbol{z}_{-j}\right) \neq y\right\}$ holds. Therefore, we know the two events in the last line of \eqref{4} are disjoint and further take the probability on both sides to get:
\begin{eqnarray}
&&P\left(Y\left(Z_j = z_j,\boldsymbol{Z}_{-j}= \boldsymbol{z}_{-j}\right)=y\mid \boldsymbol{X}= \boldsymbol{x}\right)\nonumber \\
=&&P\left(Y\left(Z_j \neq z_j,\boldsymbol{Z}_{-j}= \boldsymbol{z}_{-j}\right)=y\mid \boldsymbol{X}= \boldsymbol{x}\right)\nonumber \\
&&+P\left(Y\left(Z_j = z_j,\boldsymbol{Z}_{-j}= \boldsymbol{z}_{-i}\right)=y, Y\left(Z_j\neq z_j, \boldsymbol{Z}_{-j}=\boldsymbol{z}_{-j}\right) \neq y \mid \boldsymbol{X}= \boldsymbol{x} \right),\nonumber \\
\end{eqnarray}
where the last term is exactly $\operatorname{CPNS}_{j}$ which we want to identify.

Finally with our ignorability assumption \eqref{ignorability} we get:
\begin{eqnarray}
&&\operatorname{CPNS}_{j}\nonumber\\
=&&P\left(Y\left(Z_j = z_j,\boldsymbol{Z}_{-j}= \boldsymbol{z}_{-j}\right)=y\mid \boldsymbol{X}= \boldsymbol{x}\right)-P\left(Y\left(Z_j \neq z_j,\boldsymbol{Z}_{-j}= \boldsymbol{z}_{-j}\right)=y\mid \boldsymbol{X}= \boldsymbol{x}\right) \nonumber\\
\stackrel{\eqref{ignorability}}{=}&&P(Y=y \mid Z_{j}=z_{j}, \boldsymbol{Z}_{-j}=\boldsymbol{z}_{-j},\boldsymbol{X}= \boldsymbol{x})-P(Y=y \mid Z_{j} \neq z_{j}, \boldsymbol{Z}_{-j}=\boldsymbol{z}_{-j},  \boldsymbol{X}= \boldsymbol{x}).
\label{identify}
\end{eqnarray}

\subsection{ Lower Bound of CPNS}

\noindent \textbf{Proof:} To find the lower bound of $\operatorname{CPNS}$, for any three events $A$, $B$, and $C$, we know that 
\begin{equation}
 P(A, B \mid C)\geq \max [0, P(A\mid C)+P(B\mid C)-1].
 \label{known}
\end{equation}
We substitute $A$ for $\left\{Y\left(Z_j = z_j,\boldsymbol{Z}_{-j}= \boldsymbol{z}_{-j}\right)=y\right\}$, $B$ for $\left\{Y\left(Z_j \neq z_j,\boldsymbol{Z}_{-j}= \boldsymbol{z}_{-j}\right)\neq y\right\}$ and $C$ for $\left\{\boldsymbol{X}=\boldsymbol{x}\right\}$. 

Also similar to \eqref{identify} with ignorability assumption \eqref{ignorability}, we can get
\begin{align}
&P(A\mid C)=P(Y\left(Z_j = z_j,\boldsymbol{Z}_{-j}= \boldsymbol{z}_{-j}\right)=y \mid \boldsymbol{X}=\boldsymbol{x})=P(Y=y \mid Z_j = z_j,\boldsymbol{Z}_{-j}= \boldsymbol{z}_{-j},\boldsymbol{X}=\boldsymbol{x}).
\label{ac}\\
&P(B\mid C)=P(Y\left(Z_j \neq z_j,\boldsymbol{Z}_{-j}= \boldsymbol{z}_{-j}\right)\neq y \mid \boldsymbol{X}=\boldsymbol{x})=P(Y\neq y \mid Z_j \neq z_j,\boldsymbol{Z}_{-j}= \boldsymbol{z}_{-j},\boldsymbol{X}=\boldsymbol{x}).
\label{bc}
\end{align}

Then combining \eqref{ac} and \eqref{bc}:
\begin{align}
&P(A\mid C)+P(B\mid C)-1 \nonumber\\
=&P(Y=y \mid Z_j = z_j,\boldsymbol{Z}_{-j}= \boldsymbol{z}_{-j},\boldsymbol{X}=\boldsymbol{x})+P(Y\neq y \mid Z_j \neq z_j,\boldsymbol{Z}_{-j}= \boldsymbol{z}_{-j},\boldsymbol{X}=\boldsymbol{x})-1 \nonumber\\
=&P(Y=y \mid Z_j = z_j,\boldsymbol{Z}_{-j}= \boldsymbol{z}_{-j},\boldsymbol{X}=\boldsymbol{x})-P(Y= y \mid Z_j \neq z_j,\boldsymbol{Z}_{-j}= \boldsymbol{z}_{-j}, \boldsymbol{X}=\boldsymbol{x}).
\label{bound}
\end{align}

Finally, the lower bound can be obtained by replacing $P(A\mid C)+P(B\mid C)-1 $ in \eqref{known} by \eqref{bound}.

\end{document}